\documentclass[preprint,12pt]{elsarticle}

\usepackage{amssymb}
\usepackage{amsmath}
\usepackage{tabularx,booktabs}

\usepackage{latexsym}

\usepackage[switch]{lineno}

\usepackage{caption}
\usepackage{subcaption}

\usepackage{bm}

\usepackage{siunitx}

\usepackage{romannum}

\usepackage{float}

\usepackage{url}
\usepackage{xcolor}
\definecolor{newcolor}{rgb}{.8,.349,.1}

\usepackage[citebordercolor=white]{hyperref}
\usepackage{cleveref}

\usepackage{appendix}

\journal{Acta Astronautica}

\begin{document}

\begin{frontmatter}



\title{A Loss Landscape Visualization Framework for Interpreting Reinforcement Learning: An ADHDP Case Study}%




\author[1]{Jingyi Liu}
\author[1]{Jian Guo\corref{cor1}}
\cortext[cor1]{Corresponding author} 
\ead{J.Guo@tudelft.nl}
\author[1]{Eberhard Gill}

\affiliation[1]{organization={Faculty of Aerospace Engineering, Delft University of Technology},
                addressline={Kluyverweg 1},
                city={Delft},
                postcode={2629 HS},
                country={The Netherlands}}

\begin{abstract}
Reinforcement learning algorithms have been widely used in dynamic and control systems. However, interpreting their internal learning behavior remains a challenge. In the authors’ previous work, a critic match loss landscape visualization method was proposed to study critic training. This study extends that method into a framework which provides a multi-perspective view of the learning dynamics, clarifying how value estimation, policy optimization, and temporal-difference (TD) signals interact during training. The proposed framework includes four complementary components: a three-dimensional reconstruction of the critic match loss surface that shows how TD targets shape the optimization geometry; an actor loss landscape under a frozen critic that reveals how the policy exploits that geometry; a trajectory combining time, Bellman error, and policy weights that indicates how updates move across the surface; and a state–TD map that identifies the state regions that drive those updates. The Action-Dependent Heuristic Dynamic Programming (ADHDP) algorithm for spacecraft attitude control is used as a case study. The framework is applied to compare several ADHDP variants and shows how training stabilizers and target updates change the optimization landscape and affect learning stability. Therefore, the proposed framework provides a systematic and interpretable tool for analyzing reinforcement learning behavior across algorithmic designs.
\end{abstract}




\end{frontmatter}

\clearpage
\pagenumbering{arabic}



\section{Introduction}

\label{chapter4_sec1}



Reinforcement learning (RL) algorithms have attracted researchers' attention in recent years. They have also been applied in dynamics and control problems. 

Reinforcement learning control algorithms can be used in systems with uncertainties. For example, in an Active Debris Removal (ADR) mission, post-capture attitude control of the combined spacecraft encounters the challenges of system uncertainties, such as uncertain inertial parameters. In this case, reinforcement learning algorithms~\citep{oestreich2021autonomous,tipaldi2022reinforcement, rafiee2024active}, which rely on the sampled data during the control process, rather than the model information, have also been used.

Although reinforcement learning has achieved success in various control tasks, it doesn't always perform as expected. RL algorithms are optimized within a training environment, which is a simulation of a specific physical system. The learned policies for one RL algorithm may not transfer effectively to new or slightly different systems~\citep{packer2018assessing}. In some cases, even a minor change in a system parameter can cause a previously well-tuned algorithm to fail~\citep{peng2018sim}. RL's performance sensitivity to the training environment motivates the need to interpret the algorithm’s performance. Understanding RL's training behavior is important for revealing the learning mechanism and thus improving performance robustness.

Recent studies have explored various approaches to interpret reinforcement learning behavior in control systems. For instance, some works visualize the state–action value function to reveal the learned control patterns~\citep{glanois2024survey}, while others extract symbolic representations of the policy ~\citep{hein2020interpretable}. Methods such as sensitivity analysis, Jacobian-based visualization, and Koopman operator approximation~\citep{rozwood2024koopman} have also been introduced to relate RL policies to interpretable dynamical structures. In the authors' previous work, a visualization method that maps the critic match loss landscape was proposed to analyze the critic behavior during the training of actor–critic reinforcement learning algorithms. However, a single loss landscape alone cannot fully capture the interactions among the modules in the RL agent. Therefore, a more comprehensive and structured visualization framework is needed to support the interpretation of training dynamics and performance variations across reinforcement learning algorithms.

To help demonstrate the interpretation framework, one RL algorithm should be chosen as the sample. The Adaptive Dynamic Programming (ADP) method was proposed to obtain nearly optimal control strategies based on reinforcement learning (RL) and actor-critic (AC) architecture, which has received considerable attention in the past decades~\cite{lewis2009reinforcement, ouyang2020neural,zheng2020reinforcement,sun2021reinforcement, zhou2020incremental}.  Among the ADP method, action-dependent heuristic dynamic programming (ADHDP), a form of Q-learning, has been been applied in post-capture control for the combined spacecraft~\citep{wei2018learning,Gao2019thesis}. A notable characteristic of ADHDP is that it maintains a simpler structure, which allows faster training and adaptation in online implementation. On this basis, ADHDP is considered here to evaluate the potential of online reinforcement learning methods for attitude control of rigid combined spacecraft with unknown inertia. 

The ADHDP algorithm has shown its usage in the work mentioned above. Yet in practice, training instablities are often observed. Standard scalar metrics such as reward or TD error do not reveal exactly why training fails. These methods don't give enough information on the training information of the algorithm. Furthermore, when the ADHDP algorithm is equipped with deep learning techniques, such as a target neural network and different loss calculation methods, the underlying training procedure is not carefully studied. As a result, the ADHDP algorithm is used as a sample to explore how an interpretation framework can be applied to interpret the training process and control behavior of the ADHDP algorithm. 

In this work, a visualization framework is proposed to interpret the training behavior of actor–critic reinforcement learning algorithms. The framework reshapes the critic match loss using the one-step temporal difference (TD) error, enabling the observation of critic convergence trends throughout training. To analyze the actor’s adaptation process, an actor loss landscape is constructed to illustrate how policy updates respond to the critic’s evaluation. A time–TD–actor weight plot is introduced to capture how the policy evolves with the change of Bellman error, while a state–TD plot identifies the state regions that contribute most to TD fluctuations. Together, these visualizations form an integrated framework for diagnosing the interaction between actor, critic, and TD error during reinforcement learning training. ADHDP variants with different deep learning techniques are selected as samples for the demonstration of the interpretation framework.

In this work, Section 4.2 introduces the visualization framework and the ADHDP variants. Section 4.3 presents the visualization results and analysis for ADHDP variants' performance using landscape and trajectory analyses. Section 4.4 is the conclusion.

\section{Method}

In this section, the visualization framework for interpreting RL algorithms is introduced. The ADHDP algorithm is also given, which is used as an object to be interpreted using the visualization framework. 

\subsection{Visualizing the loss function }
\label{chapter4_subsection: visualizing the loss function}
In reinforcement learning, neural networks are employed to approximate functions, using a large number of parameters. The loss landscape, describing how the loss changes with respect to these parameters, provides insights into characteristics such as flatness, sharpness, local minima, and saddle points. Yet, the high dimensionality of the parameter space renders direct visualization intractable. To overcome this issue, the Contour Plots and Random Directions method has been introduced ~\citep{li2018visualizing}. This approach selects two directions, $\bm{\delta}$ and $\bm{\eta}$, and evaluates the loss across the 2-D plane they span. The resulting visualization corresponds to a projected slice of the overall loss landscape. The loss on this plane is defined as
\begin{equation}
f(\alpha, \beta) = \mathcal{L}(\bm{\zeta}^* + \alpha \bm{\delta} + \beta \bm{\eta}),
\label{chapter4_eq:contour-plots}
\end{equation}

where $\bm{\zeta}^*$ denotes the reference point in the parameter space, and $\alpha$ and $\beta$ are the step sizes along the two selected directions. The function $\mathcal{L}$ represents the loss value evaluated at each displaced position. This formulation transforms the high-dimensional optimization landscape into a three-dimensional surface, where $\bm{\delta}$ and $\bm{\eta}$ define the two axes, $(\alpha, \beta)$ are the coordinates in this plane,  and $f(\alpha, \beta)$ represents the vertical coordinate, thereby providing a clear and intuitive visualization of the loss landscape geometry.

\subsection{Visualization technique framework}
\label{chapter4_subsection: Visualization Technique Framework}
To give more insight into the performance of the ADHDP algorithm, more visualization figures are needed, besides the critic loss landscape visualization method. Here, the visualization technique framework based on \autoref{chapter4_subsection: visualizing the loss function} is introduced. 

To utilize the loss function visualization technique, four elements are needed for visualization, which are the recorded parameters, visualization indexes, batch data, and labels. The selection of these four elements will determine the meaning of the generated figures. In this work, four indices will be chosen to visualize the landscape, as shown in \autoref{chapter4_tab:visualization_indices}

\begin{table}[!ht]
\centering
\small 
\setlength{\abovecaptionskip}{0pt}
\setlength{\belowcaptionskip}{0pt}
\caption{Summary of visualization indices in the proposed framework.}
\label{chapter4_tab:visualization_indices}
\renewcommand{\arraystretch}{1.1}
\begin{tabularx}{\textwidth}{
    >{\raggedright\arraybackslash}p{2.6cm}
    >{\raggedright\arraybackslash}p{3.0cm}
    >{\raggedright\arraybackslash}p{3.0cm}
    >{\raggedright\arraybackslash}X}
\toprule
\textbf{Visualization index} & \textbf{Input data} & \textbf{Output figure} & \textbf{Interpretation focus} \\
\midrule
Critic match loss landscape & Recorded critic weights; states-action samples and TD targets during training & 3-D loss surface over critic weight space & Evolution of critic fitting behavior during training \\

Actor loss landscape & Recorded actor weights; frozen critic as cost function & 3-D loss surface over actor weight space & Geometry and quality of the learned policy \\

Time–TD–actor weight trajectory & Actor weights and TD errors recorded across training steps & 3-D trajectory of actor weight vs.\ TD vs.\ time & Coupling between policy drift and Bellman error \\

State–TD plot & System states and TD errors recorded during simulation & 2-D plot of state vs.\ TD & State regions that dominate TD fluctuations \\
\bottomrule
\end{tabularx}
\end{table}

The four visualization indices involve multidimensional parameters that cannot be directly plotted. To address this, a general dimensionality-reduction method is recalled. The parameters recorded during training are collected into a dataset, and Principal Component Analysis (PCA) is applied to extract the dominant directions of variation. The first one or two principal components directions, which correspond to $\bm{\delta}$ and $\bm{\eta}$ in \autoref{chapter4_eq:contour-plots}, are then used to form a low-dimensional plane or axis for visualization. This allows the multidimensional parameter updates to be represented clearly before introducing the four specific visualization techniques. For each grid on the PCA plane, its corresponding weight is calculated using

\begin{equation}
\tilde{\mathbf{w_c}}(\alpha,\beta)
=
\mathbf{w_c}^{\,r}
+
\alpha \bm{\delta}
+
\beta \bm{\eta},
\end{equation}
where $\mathbf{w_c}^{\,r}$ denotes the reference critic weight vector, $\bm{\delta}$ and $ \bm{\eta}$ are the two principal directions obtained from PCA of the recorded parameter trajectory. The loss value is evaluated on a grid over the coefficients $(\alpha,\beta)$.

The first index to be visualized is the critic match loss landscape. The critic match loss landscape will show the trend of how the critic loss landscape changes during the training process. The critic weight at the end of each episode of training is recorded. The states-action samples and corresponding TD target recorded throughout training is selected for the calculation of the loss of each grid on the landscape. 
\begin{equation}
L_{\mathrm{match}}(\alpha,\beta)
=
\frac{1}{N_{\mathrm{ref}}}
\sum_{t=1}^{N_{\mathrm{ref}}}
\left(
J(x_t^{\mathrm{state}},u_t^{\mathrm{action}};\tilde{\mathbf{w}}_c(\alpha,\beta))
-
y_t^{\mathrm{td}}
\right)^2 ,
\end{equation}
where $J(\cdot;\tilde{\mathbf{w}}_c(\alpha,\beta))$ denotes the critic network output under the reconstructed critic weight parameters, $y_t^{\mathrm{td}}$ is the TD target.

The second index to be visualized is the actor loss landscape. To fully study the training process of the RL algorithm, only showing the trend of how the critic loss changes is not enough. The actor is also approximated using neural networks. And it uses the information from the critic network as guidance. So the training of an actor neural network is also essential to evaluate the performance of the RL algorithm. With the critic frozen at the end of training as a surrogate for the cost $J(x,u)$, we define the actor loss over a fixed probe set of states $\mathcal{D}$ as
\begin{equation}
\mathcal{L}_{\text{actor}}(\bm{w}_a)
= \frac{1}{|\mathcal{D}|}\sum_{x\in\mathcal{D}} \,\hat J\!\big(x,\,\pi_{\bm{w}_a}(x)\big),
\end{equation}
where $\hat J(x,u)$ is the critic output and $\mathcal{D}$ is a fixed set of probe states collected during training.We record the actor parameters $\mathbf{w}_a$ at the end of each episode. To visualize the landscape, we evaluate $\mathcal{L}_{\text{actor}}(\theta)$ on a 2-D grid in parameter space constructed around the recorded trajectory, which is the PCA plane of $\{\mathbf{w}_a\}$, yielding an actor loss surface.

The third index to be visualized is the update of the actor weight with time and TD error. This index also shows how the actor's weight changes with TD error. The actor's weight is recorded after several training steps, and the corresponding TD error is recorded. Since the actor's weight is a multi-dimensional parameter, the PCA technique is applied. The first main direction of the weight after PCA is selected for the visualization, together with time and TD error.

The last index to be visualized is the update of the state trajectory with TD error. Similar to how the third index is visualized, the states are recorded per several simulation steps, the corresponding TD error is recorded. PCA is also applied to the multidimensional states of the dynamic system. Then the states and TD error can be plotted in the 2-D figure. This visualization helps identify regions of the state space that contribute most to large TD errors.

\subsection{ADHDP variants with deep learning techniques}
\begin{figure}[htbp]
\centering
\includegraphics[width = 0.7\textwidth]{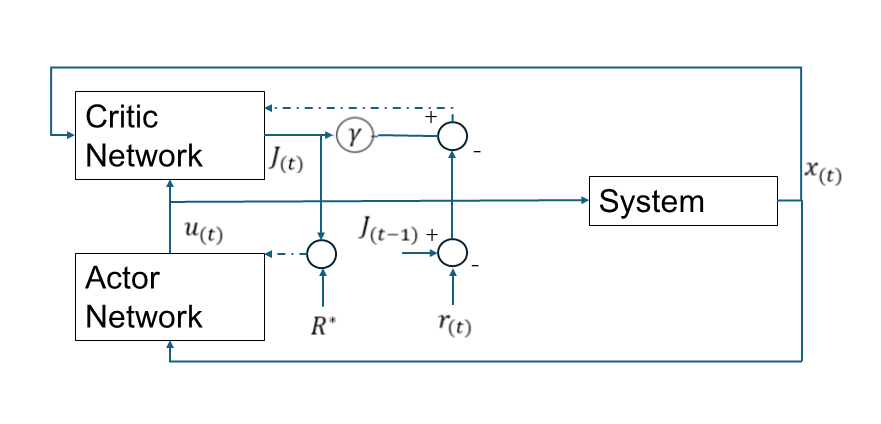}
\caption{Structure of ADHDP}
\label{chapter4_fig:ADHDP_structure}
\end{figure}

To demonstrate how visualizing the critic loss landscape can aid in interpreting online RL algorithms, we take Action-dependent Heuristic Dynamic Programming (ADHDP) as a representative example. ADHDP is a particular reinforcement learning framework in which the control and learning objectives are separated and handled by two distinct neural networks: the actor and the critic, as illustrated in ~\autoref{chapter4_fig:ADHDP_structure} ~\citep{si2004handbook}. The critic network is optimized to approximate the total reward-to-go, serving as a surrogate for the cost function, while the actor network is trained so that the critic ultimately drives the system toward minimizing this cost. 

In ADHDP, the cost function $J(t)$ is formulated as
\begin{eqnarray}
J(t)=r(t+1)+\gamma r(t+2)+\cdots=\sum_{k=1}^{\infty} \gamma^{k-1} r(t+k),
\label{chapter4_eq:discounted cost function}
\end{eqnarray}
where $r(t+1)$ denotes the reinforcement signal (or reward) received by the system, and $k$ indicates the time steps from instant $t$. The discount factor $\gamma$ determines the relative importance of future rewards in the cumulative return.  

The prediction error for the critic network, denoted by $e_c$, is defined as
\begin{eqnarray}
e_c(t) = [r(t) + \gamma J(t)] - J(t-1),
\label{chapter4_eq:TD Error}
\end{eqnarray}
which corresponds to the temporal-difference (TD) error commonly used in reinforcement learning.  

Let $\mathbf{w}_c$ represent the weight parameters of the critic network. Their update follows the gradient descent principle. In this study, both the critic and actor networks are implemented as multilayer perceptrons (MLPs).

\autoref{chapter4_fig:ADHDP_structure} and \autoref{chapter4_eq:TD Error} describe the core of the ADHDP algorithm. The performance of the ADHDP algorithm will largely depend on the reinforcement learning techniques employed in the ADHDP algorithm. In this work, four versions of the ADHDP algorithm are interpreted using the visualization framework in \autoref{chapter4_subsection: Visualization Technique Framework}. The deep learning techniques employed by each version is shown in \autoref{chapter4_tab:adhdp_variants}.

\begin{table}[!ht]
\centering
\small
\setlength{\abovecaptionskip}{0pt}
\setlength{\belowcaptionskip}{0pt}
\caption{DRL techniques applied in each ADHDP variant. The numbering corresponds to the order of variants discussed in the text.}
\label{chapter4_tab:adhdp_variants}
\renewcommand{\arraystretch}{1.1}
\begin{tabularx}{\textwidth}{
>{\centering\arraybackslash}p{0.5cm}
>{\raggedright\arraybackslash}p{2.5cm}
>{\centering\arraybackslash}p{1.8cm}
>{\centering\arraybackslash}p{1.8cm}
>{\centering\arraybackslash}p{1.8cm}
>{\centering\arraybackslash}p{2.2cm}}
\toprule
\textbf{No.} & \textbf{ADHDP variant} & \textbf{Target network} & \textbf{Numeric cost scaling} & \textbf{Huber critic loss} & \textbf{Target-policy smoothing (TPS)} \\
\midrule
1 & Basic version & -- & -- & -- & -- \\
2 & With target network & \checkmark & -- & -- & -- \\
3 & With training stabilizers + TPS & \checkmark & \checkmark & \checkmark & \checkmark \\
4 & With training stabilizers only & \checkmark & \checkmark & \checkmark & -- \\
\bottomrule
\end{tabularx}
\end{table}

The simplified version corresponds to the basic ADHDP algorithm described in \autoref{chapter4_fig:ADHDP_structure}. The second version introduces a target network for the critic. The third version builds on the target-network version and further incorporates two training stabilizers, namely numeric cost scaling and the Huber critic loss. Target policy smoothing is also applied in this version. The fourth version removes the target policy smoothing used in the third version while retaining the two training stabilizers (Huber critic loss and numeric cost scaling).

Below, the DRL techniques employed in the ADHDP variants are introduced. 

\paragraph{Target network}
Taking the critic network as an example, the target network is used to provide a less chaotic update to the critic network. A soft update ,the Polyak averaging, of the target network is applied in this work. After each gradient step on the online critic, the target critic is softly updated toward it:

\begin{equation}
\bar{\mathbf{w}_c} \leftarrow (1 - \tau_{\text{target}})\,\bar{\mathbf{w}_c} + \tau_{\text{target}}\,\mathbf{w}_c,
\end{equation}

with a small $\tau_{\text{target}} \in (0,1)$. Smaller $\tau_{\text{target}}$ yields a slower, more stable update.

\paragraph{Numeric cost scaling~\citep{henderson2018deep}}
This technique is defined as one of the two training stabilizers in this work. 
We rescale the per-step cost by a positive constant scaling factor 
$\alpha_{\text{cost}} > 0$:
\begin{equation}
 c_t' = \tfrac{1}{\alpha_{\text{cost}}}\, c_t, 
\qquad 
J'(x) = \tfrac{1}{\alpha_{\text{cost}}}\, J(x).
\end{equation}
Here $c_t$ denotes the original instantaneous cost at time step $t$, $c_t'$ is the scaled cost used in training, $J(x)$ represents the value function associated with state $x$, and $J'(x)$ is the corresponding value function under the scaled cost.

Since this is a monotone linear transform, the optimal policy is unchanged. Only the numerical range of TD targets and gradients is compressed, which reduces loss curvature and gradient explosions.  

\paragraph{Huber critic loss}
This technique is defined as one of the two training stabilizers in this work. 
Critic parameters $\mathbf{w}_c$ minimize a robust regression between prediction and target:

\begin{equation}
L_{\text{critic}} =
\begin{cases}
    \tfrac{1}{2}\, e^2, & \lvert e \rvert \leq \kappa_{\text{huber}}, \\[2mm]
    \kappa_{\text{huber}} \lvert e \rvert - \tfrac{1}{2}\kappa_{\text{huber}}^2, & \lvert e \rvert > \kappa_{\text{huber}},
\end{cases}
\quad
e = \hat{J}_{\mathbf{w}_c}(x_t, u_t) - y_t^{\mathrm{td}},
\label{eq:huber_loss}
\end{equation}

where the symbols are defined as follows:
$y_t^{\mathrm{td}}$ is the target value for the critic at time step $t$, typically obtained via bootstrapped estimates or TD targets. $\kappa_{\text{huber}}$ is the threshold parameter in the Huber loss that determines the transition from quadratic ($\ell_2$) to linear ($\ell_1$) behavior. $e$ is the prediction error of the critic, i.e., the difference between the predicted value $\hat{J}_{\mathbf{w}_c}(x_t, u_t)$ and the target $y_t^{\mathrm{td}}$.

Near zero, the Huber loss behaves like the $\ell_2$ loss, promoting smooth convergence, 
while in the tails it behaves like the $\ell_1$ loss, down-weighting outliers arising from bootstrapping or distribution shifts.
\paragraph{Target-policy smoothing (TPS)~\citep{fujimoto2018addressing}} 
When forming the TD target, we evaluate the next action using a noisy, clipped version of the policy, then feed it to the target critic:
\begin{equation}
\tilde{u}_{t+1} =
\operatorname{clip}\Big(
\mu_{\theta}(x_{t+1}) + \epsilon_{\text{noise}},
\, -u_{\max},\, u_{\max}
\Big),
\qquad
\epsilon_{\text{noise}} \sim
\mathcal{N}\big(0, \sigma_{\text{noise}}^{2}\mathbf{I}\big)
\end{equation}

The TD target is
\begin{equation}
y_t^{\mathrm{td}} =
\begin{cases}
c_t' + \gamma \,\hat{J}_{\bar{\phi}}\!\left(x_{t+1}, \tilde{u}_{t+1}\right), & \text{if not terminal}, \\[6pt]
c_t' + \text{penalty}', & \text{terminal}.
\end{cases}
\end{equation}

Here $\hat{J}_{\bar{\phi}}$ denotes the slow-moving target critic network obtained via soft updates. The term $\epsilon_{\text{noise}}$ represents the policy smoothing noise added to the next action, which is sampled from a zero-mean Gaussian distribution with standard deviation $\sigma_{\text{noise}}$. The parameter $\sigma_{\text{noise}}$ controls the magnitude of the injected noise. $\mathbf{I}$ denotes the identity matrix, ensuring that the noise is applied independently to each action dimension. $c_t'$ is the scaled instantaneous cost used for critic training, obtained from the original cost $c_t$ through the numeric cost scaling described previously. The clipping operation enforces the actuator bounds defined by the maximum control torque $u_{\max}$.

TPS acts as a local averaging operator over the action dimension, reducing over-estimation bias and spiky curvature in $J(x,u)$ around the greedy action.

\section{Results for demonstration of the framework}

In this section, the spacecraft attitude system is introduced. The control results of using different versions of the ADHDP algorithm are shown. The control results will be interpreted using the loss landscape visualization framework.

\subsection{Attitude dynamics of spacecraft with unknown inertia}
\label{chapter4_attitude dynamics}

In this study, the combined spacecraft in the ADR scenario is taken as the control system for testing ADHDP variants. As reviewed in \autoref{chapter4_sec1}, the combined spacecraft in an ADR mission is subject to multiple sources of uncertainty. To evaluate the control algorithm for the post-capture phase in a progressive manner, we begin with the simplest case, where uncertainty arises solely from the inertial parameters. The following assumptions are made. 
\begin{itemize}
  \item The target is firmly grasped by rigid robotic manipulators mounted on a rigid servicing spacecraft.
  \item After capture, all manipulator joints are locked ~\citep{huang2018postcapture}.
  \item The target is rigid, uncooperative, and has no control capability.
\end{itemize}

Under these conditions, the combined spacecraft can be modeled as a single rigid body with unknown inertial properties.  

In this work, the reinforcement learning algorithm is trained in simulation instead of on the physical spacecraft. The training process produces an initial stabilizing control policy represented by the actor network. During operation, the learned policy is used as a feedback controller that generates control inputs from the observed spacecraft states. Training in simulation improves safety, since the real system does not need to start from an untrained policy. In addition, the visualization framework can be used during training to examine the evolution of the critic and actor optimization, providing insight into the learning process before deployment.

A body-fixed reference frame $\mathcal{B}$ is defined at the center of mass of the combined spacecraft, with its axes aligned to the principal inertia directions. An Earth-Centered Inertial (ECI) frame $\mathcal{N}$ is also introduced, spanned by the unit vectors $\{\bm{\overline{n_1}}, \bm{\overline{n_2}}, \bm{\overline{n_3}}\}$. The $\bm{\overline{n_1}}$ axis points toward the mean equinox, $\bm{\overline{n_3}}$ is aligned with the Earth’s rotation axis (celestial north), and $\bm{\overline{n_2}}$ completes the right-handed triad.  

The spacecraft attitude is represented by a unit quaternion  
\[
\bm{q} = 
\begin{bmatrix}
q_0 \\
q_1 \\
q_2 \\
q_3
\end{bmatrix},
\]
which specifies the rotation from $\mathcal{N}$ to $\mathcal{B}$. This formulation avoids singularities and is well-suited for large-angle maneuvers. The angular velocity, expressed in frame $\mathcal{B}$, is denoted as  
\[
\bm{\omega} =
\begin{bmatrix}
\omega_1 \\
\omega_2 \\
\omega_3
\end{bmatrix}.
\]

The kinematic equation of motion is
\begin{equation}
\dot{\bm{q}} = \frac{1}{2} \, \bm{q} \otimes 
\begin{bmatrix}
0 \\
\bm{\omega}
\end{bmatrix},
\label{chapter4_eq:sc_kinematics}
\end{equation}
where $\otimes$ denotes quaternion multiplication.  

The attitude dynamics are governed by
\begin{equation}
\bm{M} = \hat{J}_{\mathrm{sc}} \cdot \dot{\bm{\omega}} + \bm{\omega} \times \left( \hat{J}_{\mathrm{sc}} \cdot \bm{\omega} \right),
\label{chapter4_eq:sc_dynamics}
\end{equation}
where $\bm{M}$ is the applied control torque, and $\hat{J}_{\mathrm{sc}}$ is the (unknown) inertia matrix of the combined spacecraft.

\subsection{Common simulation setup}

The following settings are shared across all ADHDP simulations. The integration step is set to $dt = 0.01\,\text{s}$, with $5000$ steps per episode and $100$ episodes in total. 

Regarding the setup of the inertia parameters of the spacecraft, in practical ADR scenarios, the inertia properties of the combined spacecraft are generally uncertain due to the unknown mass distribution of the captured target. In this work, the control problem is therefore formulated as the stabilization of a rigid spacecraft with unknown inertia parameters. The reinforcement learning algorithms considered in this study are model-free, meaning that the controller does not require knowledge of the inertia matrix or other system parameters. Instead, the control policy is learned through interaction with the system dynamics. For simulation purposes, however, the spacecraft dynamics must be generated using a specific inertia matrix. Therefore, a fixed inertia value is used in the numerical simulations to represent the dynamics of the combined spacecraft. This inertia value is not provided to the learning algorithm and is only used within the simulation environment to produce state transitions.
The spacecraft inertia matrix is
\[
J = 
\begin{bmatrix}
1.0 & 0.02 & 0.02 \\
0.02 & 0.8 & 0.03 \\
0.02 & 0.03 & 0.9
\end{bmatrix},
\]
and no external disturbance torque is applied. 
The maximum control torque is limited to $0.5\,\text{N·m}$. 
The initial attitude corresponds to a $20^\circ$ rotation about the axis $[1.0,\,-0.5,\,0.2]$, with zero initial angular velocity. 
The discount factor is $\gamma = 0.95$. 
Both actor and critic are two-layer MLPs with $64$ hidden units, trained with learning rates of $1\times10^{-4}$ and $1\times10^{-3}$, respectively. 
Gaussian policy noise decays from $0.02$ to $0.005$ over $50{,}000$ steps. 
The cost are defined using \autoref{chapter4_eq:s/c_cost}, with weights coefficients $k_{\mathrm{att}} = 20.0$, $k_{\mathrm{rate}} = 2.0$, and $k_{\mathrm{torque}} = 0.1$. 
Gradients are clipped at $1.0$, and a termination penalty of $300$ is applied if the attitude error exceeds $2.8\,\text{rad}$ or the angular velocity norm exceeds $8\,\text{rad/s}$.

\begin{equation}
c_t =
k_{\mathrm{att}}\left(1-q_0^2\right)
+
k_{\mathrm{rate}}\,\|\bm{\omega}\|^2
+
k_{\mathrm{torque}}\,\|\bm{M}\|^2 
\label{chapter4_eq:s/c_cost}
\end{equation}

\subsection{Basic ADHDP}
Under the given common simulation setup above, the simulation results for spacecraft attitude control using the basic ADHDP algorithm are shown below. 
\begin{figure}[htbp]
     \centering
     \begin{subfigure}[b]{0.55\textwidth}
         \centering
         \includegraphics[width=\textwidth]{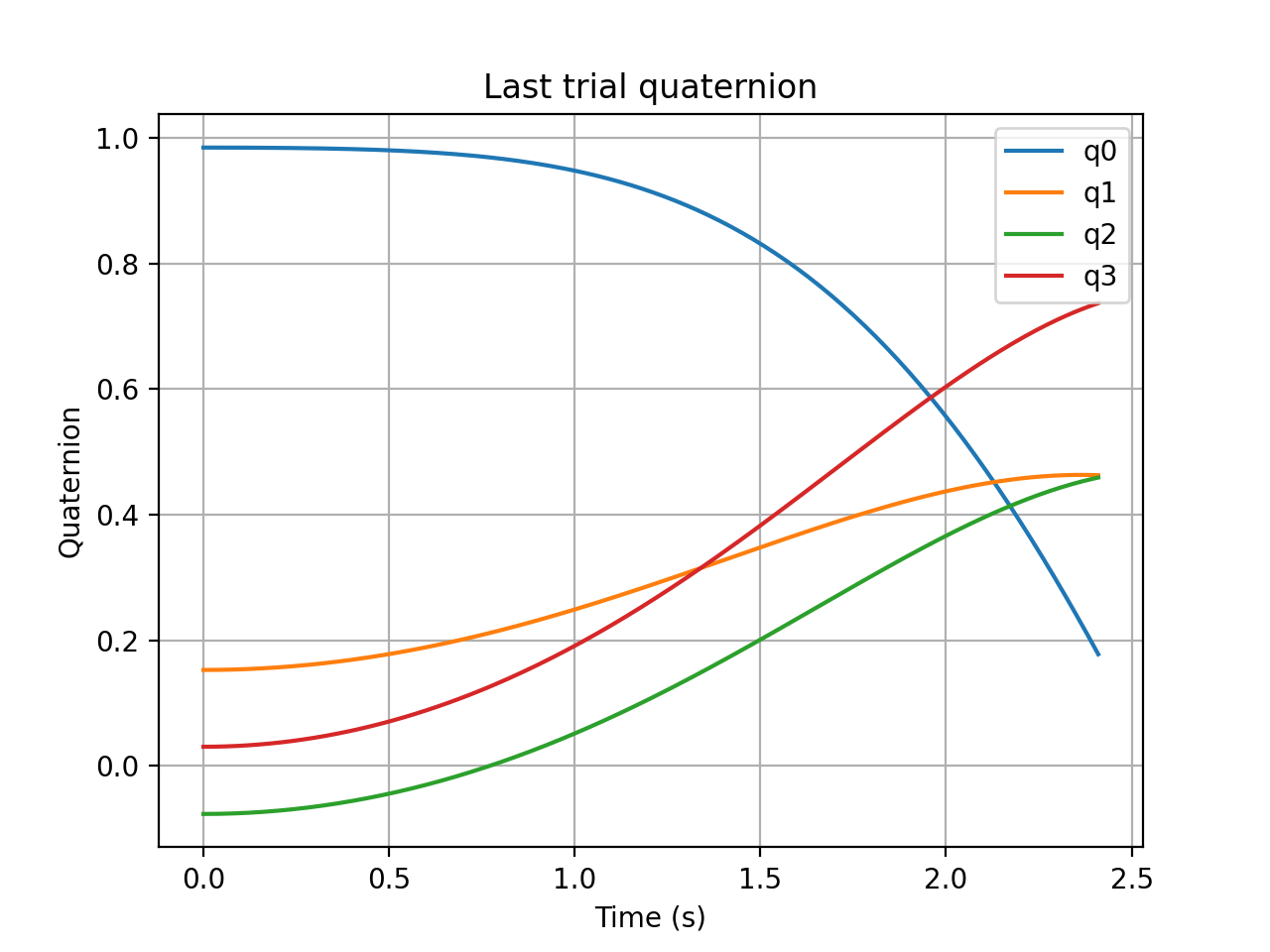}
         \caption{Quaternions}
         \label{chapter4_fig:basicADHDPquat}
     \end{subfigure}
     \hfill
     \begin{subfigure}[b]{0.55\textwidth}
         \centering
         \includegraphics[width=\textwidth]{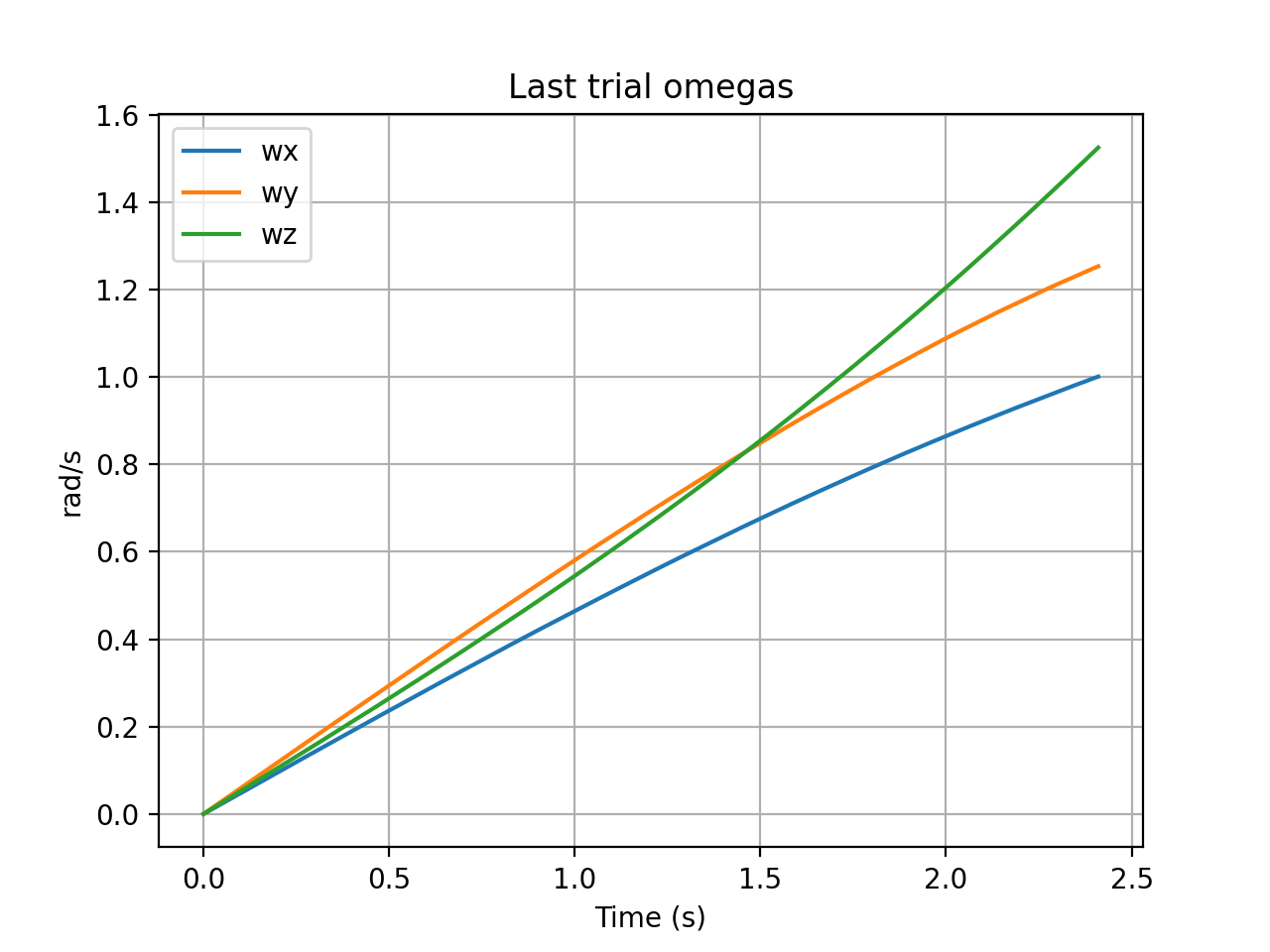}
         \caption{Angular velocity}
         \label{chapter4_fig:basicADHDPomega}
     \end{subfigure}
     \hfill
     \begin{subfigure}[b]{0.55\textwidth}
         \centering
         \includegraphics[width=\textwidth]{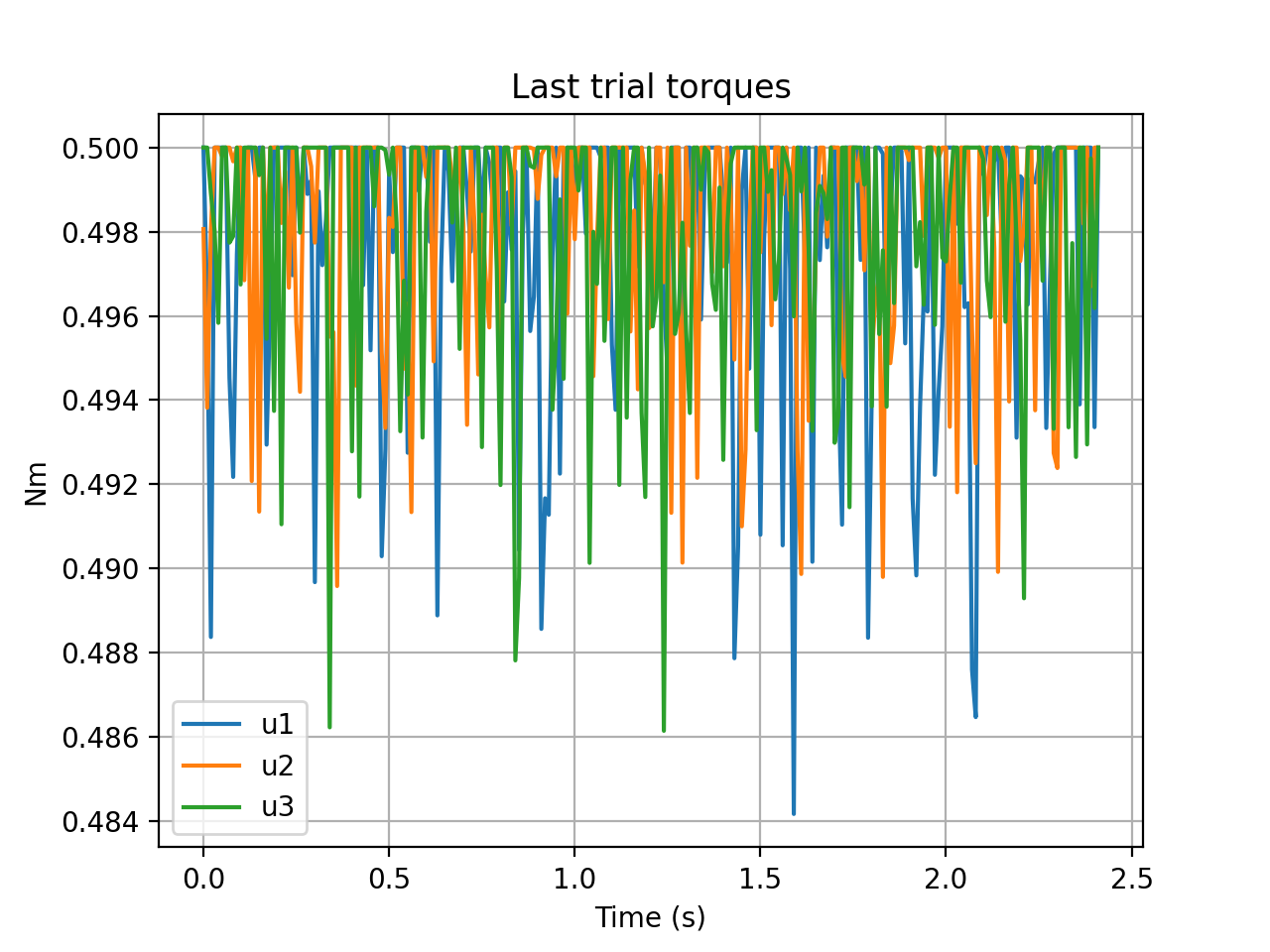}
         \caption{Control Torque}
         \label{chapter4_fig:basicADHDPtorque}
     \end{subfigure}
        \caption{State trajectory, basic ADHDP}
        \label{chapter4_fig:basicADHDPstates}
\end{figure}

With the basic version of the ADHDP algorithm, \autoref{chapter4_fig:basicADHDPstates} shows the quaternions, angular velocities, and control torque of the system during the last episode of training. It shows the failed control result. The torques of the two axes are almost saturated. With the figures below, the reasons for failure will be given.

\begin{figure}[htbp]
     \centering
     \begin{subfigure}[b]{0.49\textwidth}
         \centering
         \includegraphics[width=\textwidth]{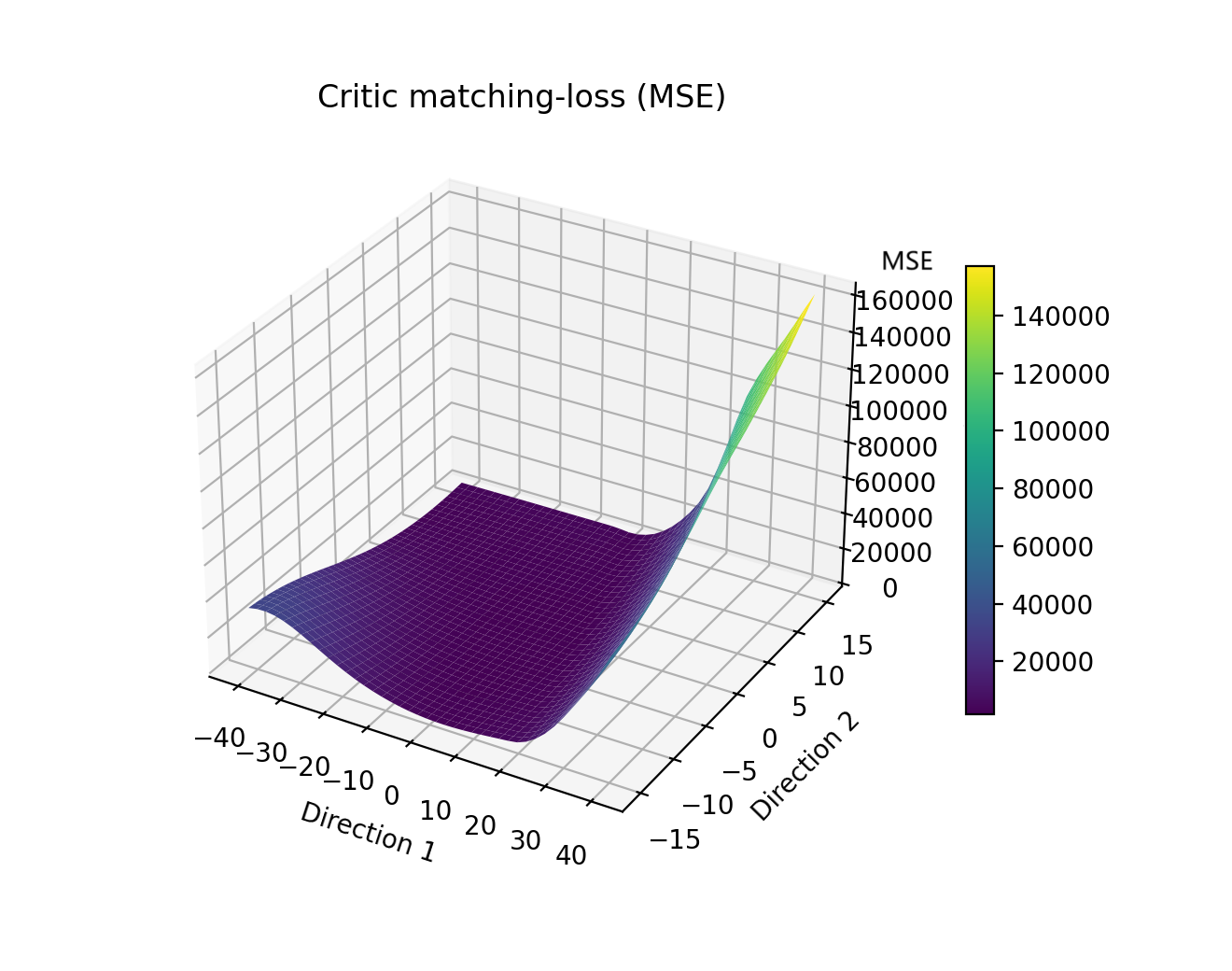}
         \caption{Critic Match Loss Landscape }
         \label{chapter4_fig:targetADHDPcritic}
     \end{subfigure}
     \begin{subfigure}[b]{0.49\textwidth}
         \centering
         \includegraphics[width=\textwidth]{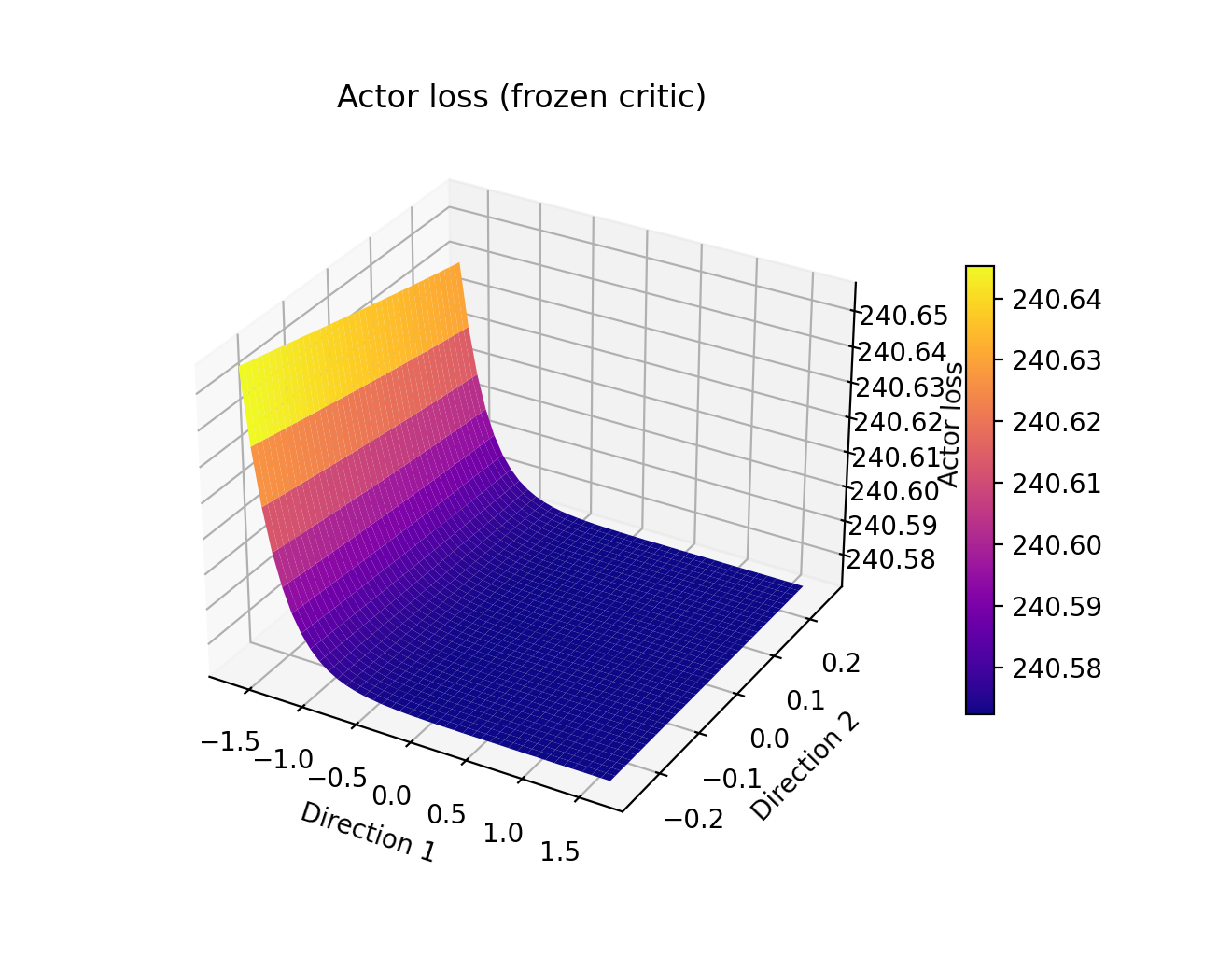}
         \caption{Actor Loss Landscape}
         \label{chapter4_fig:basicADHDPactorloss}
     \end{subfigure}
        \caption{Loss surface, basic ADHDP}
        \label{chapter4_fig:basicADHDPlosssurface}
\end{figure}

\autoref{chapter4_fig:basicADHDPlosssurface} shows the loss surface of the basic version of ADHDP during training. From \autoref{chapter4_fig:targetADHDPcritic} we can see that the critic match loss is sharp in one direction yet quite flat in another direction. \autoref{chapter4_fig:basicADHDPactorloss} shows the actor loss landscape under the final critic network. The surface is flat in the first direction, yet it shows a much smaller range of change in the second direction. 
It indicates that, under the current critic guidance, a large region of actions gives equal solutions. This flat valley makes the policy gradient easy to stop near saturation. 

\begin{figure}[htbp]
     \centering
     \begin{subfigure}[b]{0.5\textwidth}
         \centering
         \includegraphics[width=\textwidth]{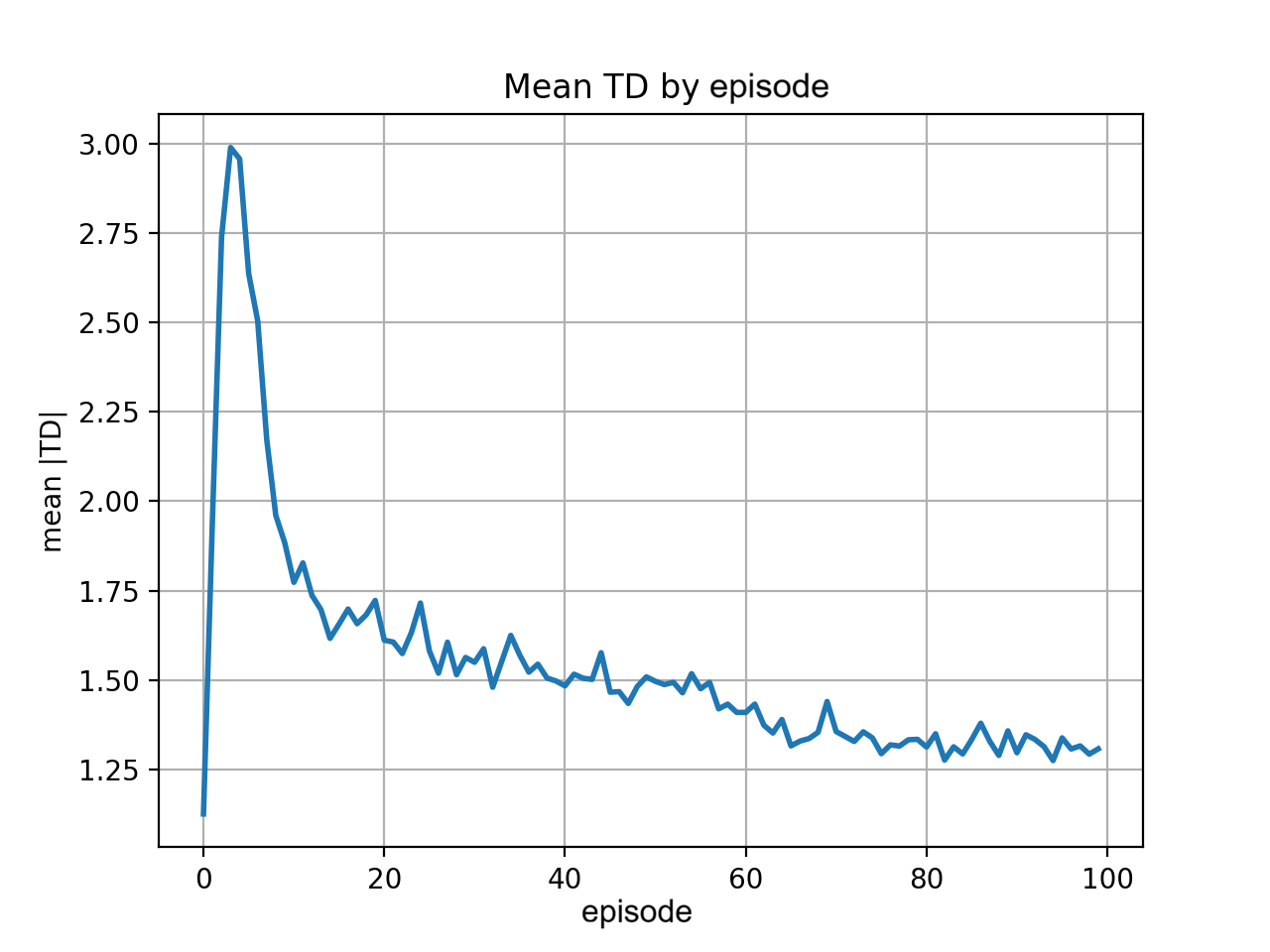}
         \caption{TD by episode}
         \label{chapter4_fig:basicADHDPTDbytrial}
     \end{subfigure}
     \hfill
     \begin{subfigure}[b]{0.65\textwidth}
         \centering
         \includegraphics[width=\textwidth]{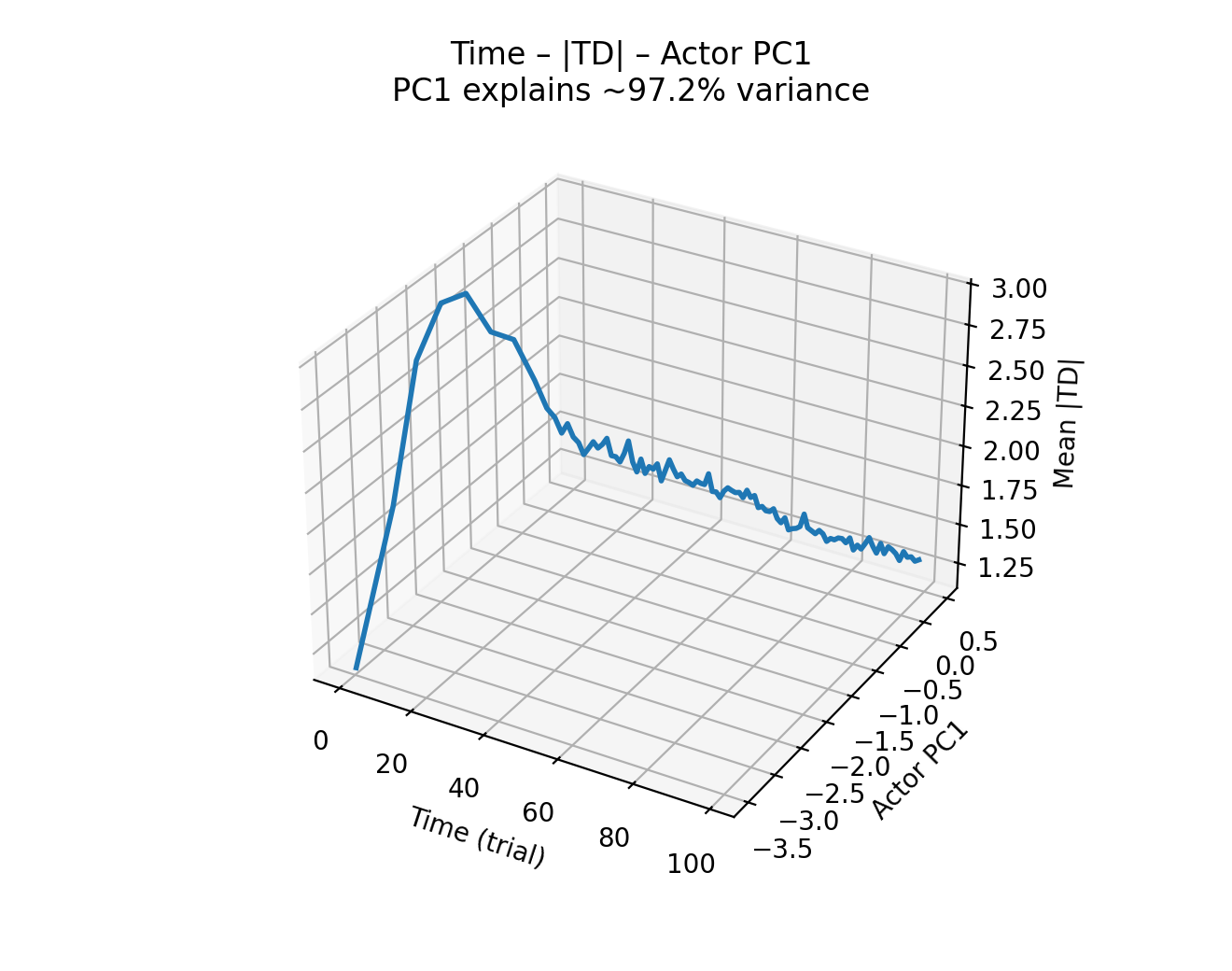}
         \caption{Actor weight with time and TD}
         \label{chapter4_fig:basicADHDPactweightTD}
     \end{subfigure}
     \hfill
     \begin{subfigure}[b]{0.5\textwidth}
         \centering
         \includegraphics[width=\textwidth]{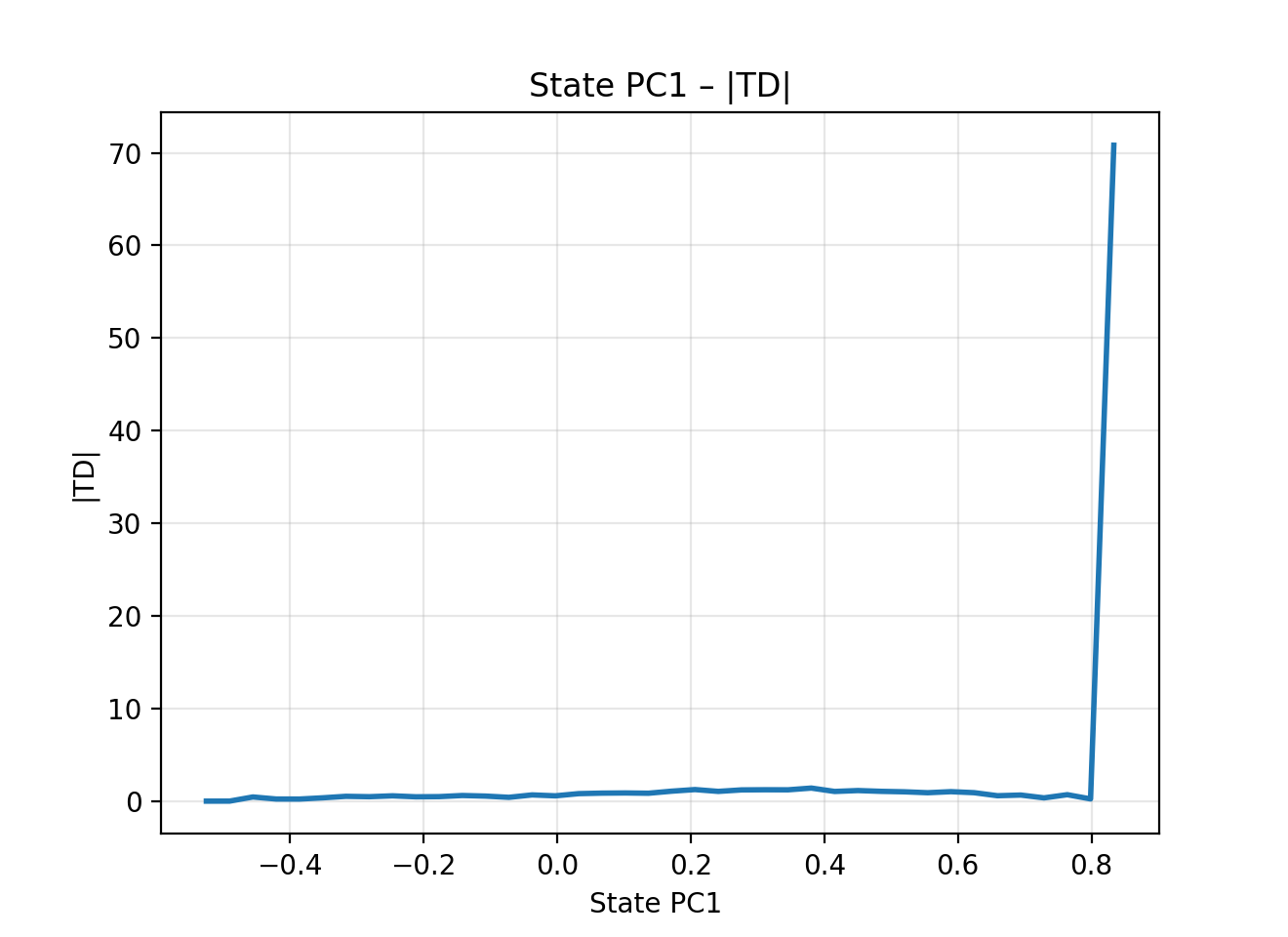}
         \caption{State with TD}
         \label{chapter4_fig:fig:basicADHDPstateTD}
     \end{subfigure}
        \caption{TD trajectory, basic ADHDP}
        \label{chapter4_fig:basicADHDPTD}
\end{figure}

\autoref{chapter4_fig:basicADHDPactweightTD} shows the mean TD of each episode during training. The TD error first rapidly increases and then slowly decreases, and finally oscillates around 1.3. It shows no convergence trend.  \autoref{chapter4_fig:basicADHDPactweightTD} shows the first principle of the actor weight change with time and TD error. It has a variance of 97.2\%, indicating that the weight evolution is almost in a single direction. The curve increases rapidly in the front part and slows down in the middle and back parts.  It is consistent with the actor surface in \autoref{chapter4_fig:basicADHDPactorloss}. The policy drifts in large steps in the early stage and is limited by the flat valley in the later stage. \autoref{chapter4_fig:fig:basicADHDPstateTD} shows how the first principle direction of states changes with time and TD. It has an extremely high TD spike at the end of the state, while the rest of the graph is relatively low. Most states exhibit relatively small TD errors, indicating that the critic approximates the value function reasonably well in the commonly visited regions of the state space. However, the extreme PC1 region shows a sharp TD spike, suggesting that the critic fails to generalize to states with large attitude errors and angular velocities. The distribution of states differs significantly from that of the early training phase, making it difficult for the critic to generalize.

From the figures above, we can conclude that the failure of this simulation is caused by the unstable update of the critic net, which leads the actor net to saturate at the boundary of the flat valley. These two factors together lead to the divergence of training.

The above analysis indicates that the failure of the basic ADHDP algorithm is closely related to the instability of the critic update. The TD error exhibits significant fluctuations during training, and the critic loss landscape shows strong anisotropy, suggesting that the critic is trained with rapidly changing targets. Since the actor update directly depends on the critic estimates, this instability propagates to the actor network and leads to saturation in the flat valley of the actor loss landscape. To improve the stability of the critic update, a target network can be introduced in the ADHDP framework. In the following section, the same diagnostic framework is applied to analyze the training behavior of ADHDP with a target network and to examine how the target network influences the critic and actor dynamics.

\subsection{ADHDP with target network}

This variant of the ADHDP algorithm incorporates a target critic network with Polyak averaging with $\tau_{\text{target}} = 0.005$ to stabilize the critic bootstrap. 
The actor and critic networks are otherwise configured as in the common setup. 
Training uses a delayed actor update. The actor is updated every 5 steps, starting after 30000 steps of pretraining freeze. Pretrain freeze is applied in training. Actor parameters are frozen for the first 10{,}000 environment steps. Critic trains normally during this phase. After pretrain freeze, actor is updated only every 10 steps. Gradient clipping $1.0$ and L2 regularization $\lambda_\text{actor} = 0.1$ are applied. Exploration noise is applied to the action. Environment action noise decays from $\sigma_\text{noise,init} = 0.05$ to $\sigma_\text{noise, final} = 0.02$ over 50{,}000 steps.

\begin{figure}[htbp]
     \centering
     \begin{subfigure}[b]{0.55\textwidth}
         \centering
         \includegraphics[width=\textwidth]{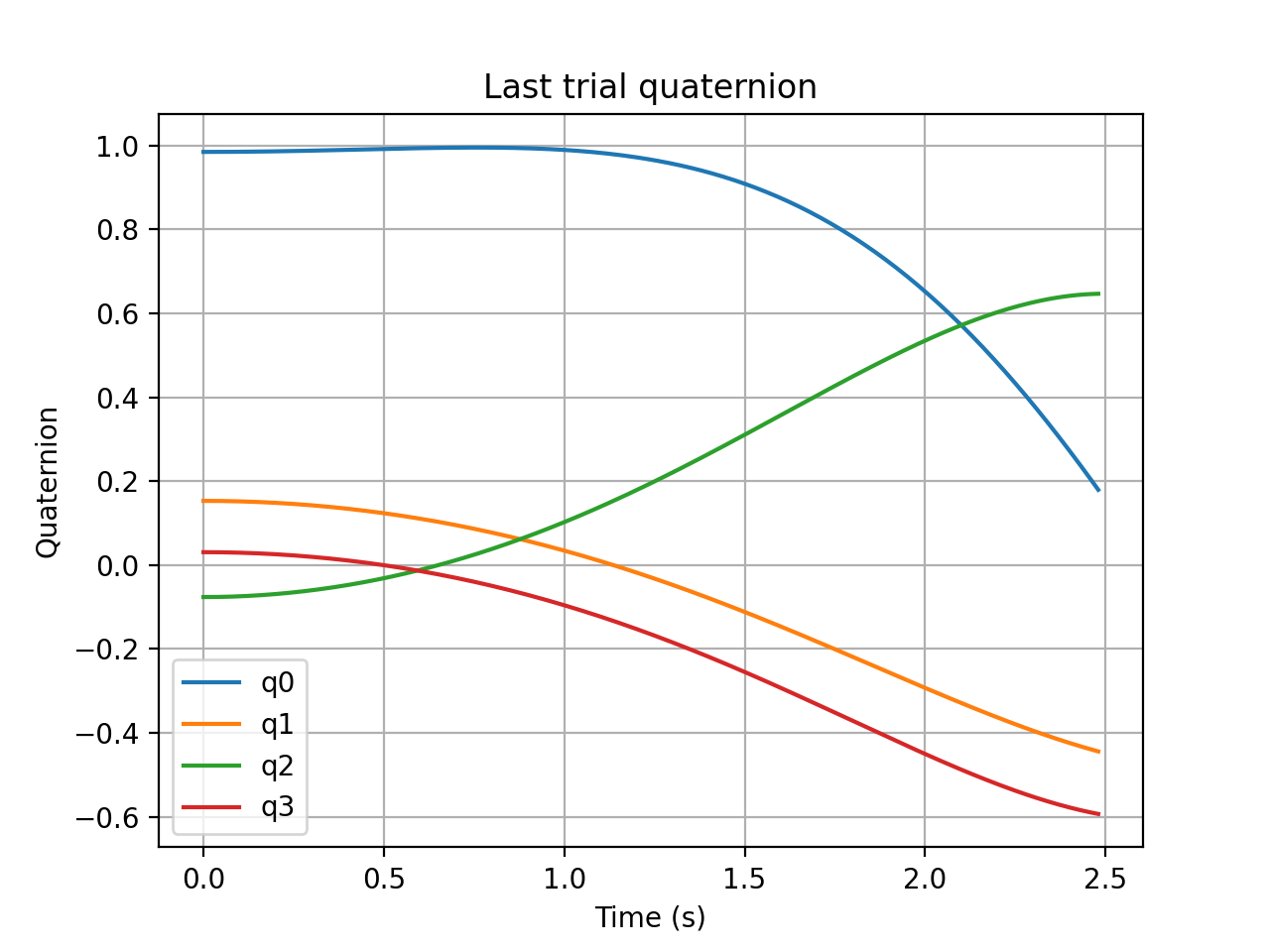}
         \caption{Quaternions}
         \label{chapter4_fig:targetADHDPquat}
     \end{subfigure}
     \hfill
     \begin{subfigure}[b]{0.55\textwidth}
         \centering
         \includegraphics[width=\textwidth]{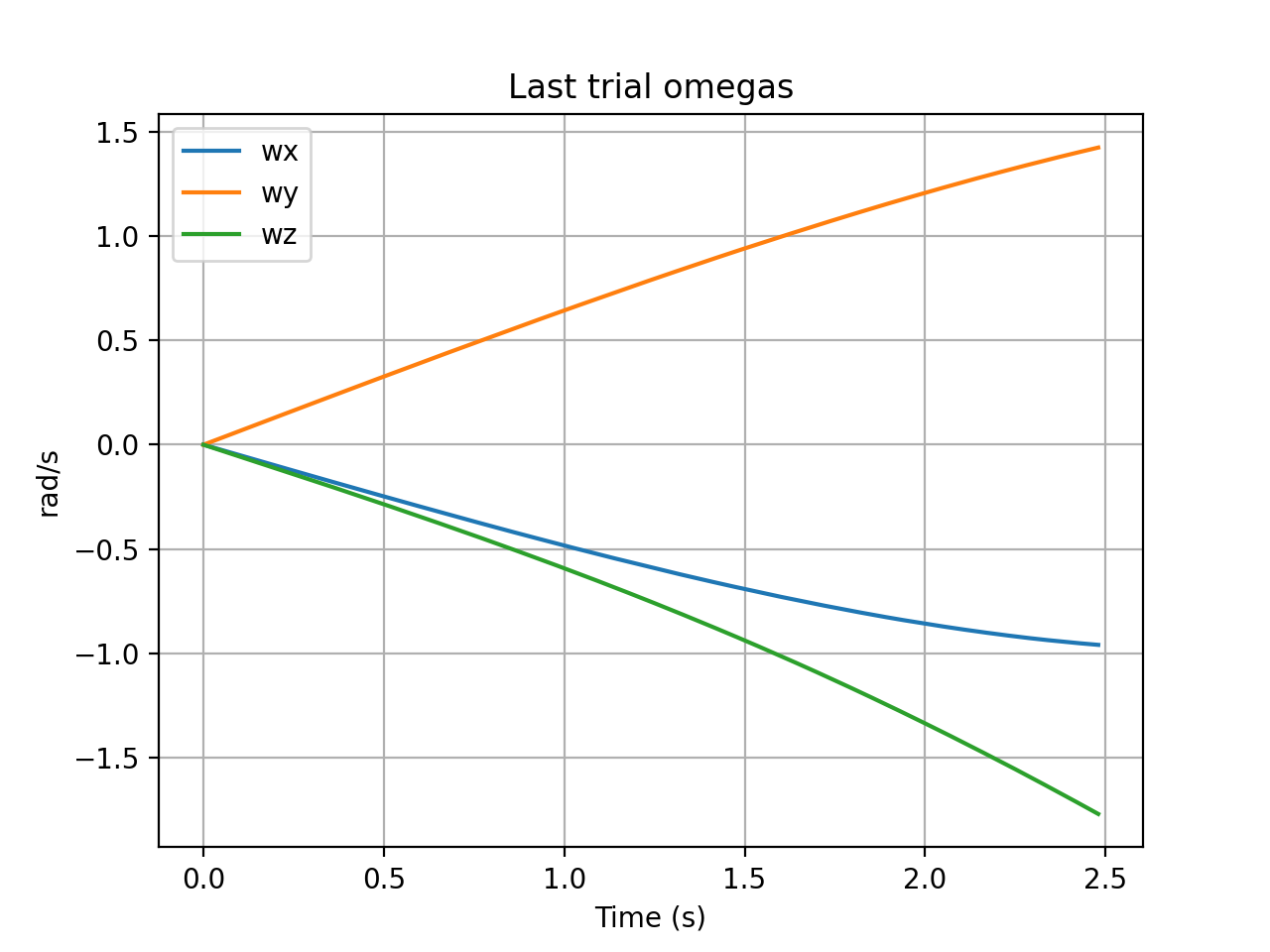}
         \caption{Angular velocity}
         \label{chapter4_fig:targetADHDPomega}
     \end{subfigure}
     \hfill
     \begin{subfigure}[b]{0.55\textwidth}
         \centering
         \includegraphics[width=\textwidth]{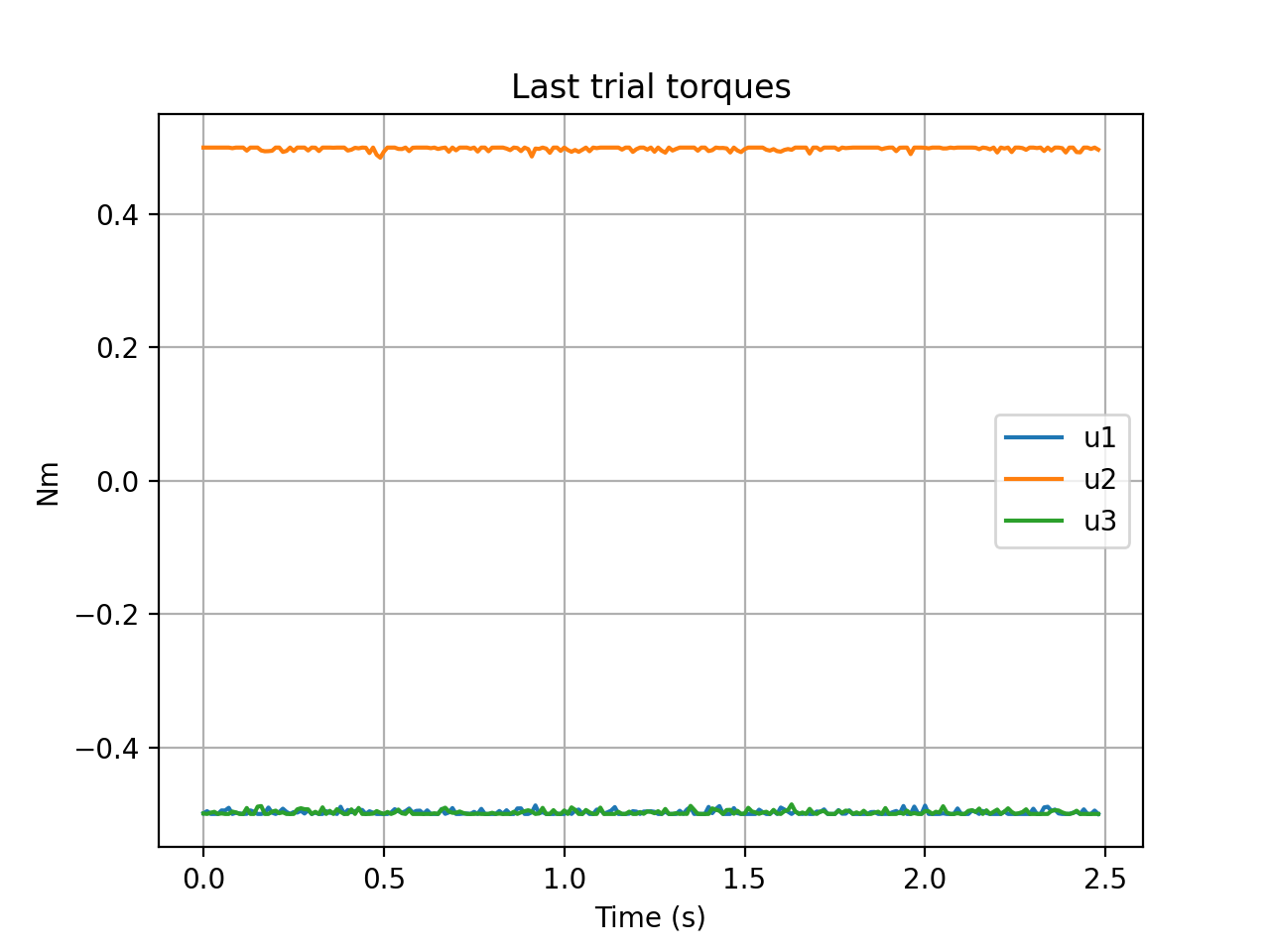}
         \caption{Control Torque}
         \label{chapter4_fig:targetADHDPtorque}
     \end{subfigure}
        \caption{State trajectory, ADHDP with target net}
        \label{chapter4_fig:targetADHDPstates}
\end{figure}
With the version of the ADHDP algorithm with target net, \autoref{chapter4_fig:targetADHDPstates} shows the quaternions, angular velocities, and control torque of the system during the last episode of training. Quaternions and attitude angular velocity trajectories are still unstable. Control toques still have obvious saturation. The reasons for the non-convergence will be given.

\begin{figure}[htbp]
     \centering
     \begin{subfigure}[b]{0.49\textwidth}
         \centering
         \includegraphics[width=\textwidth]{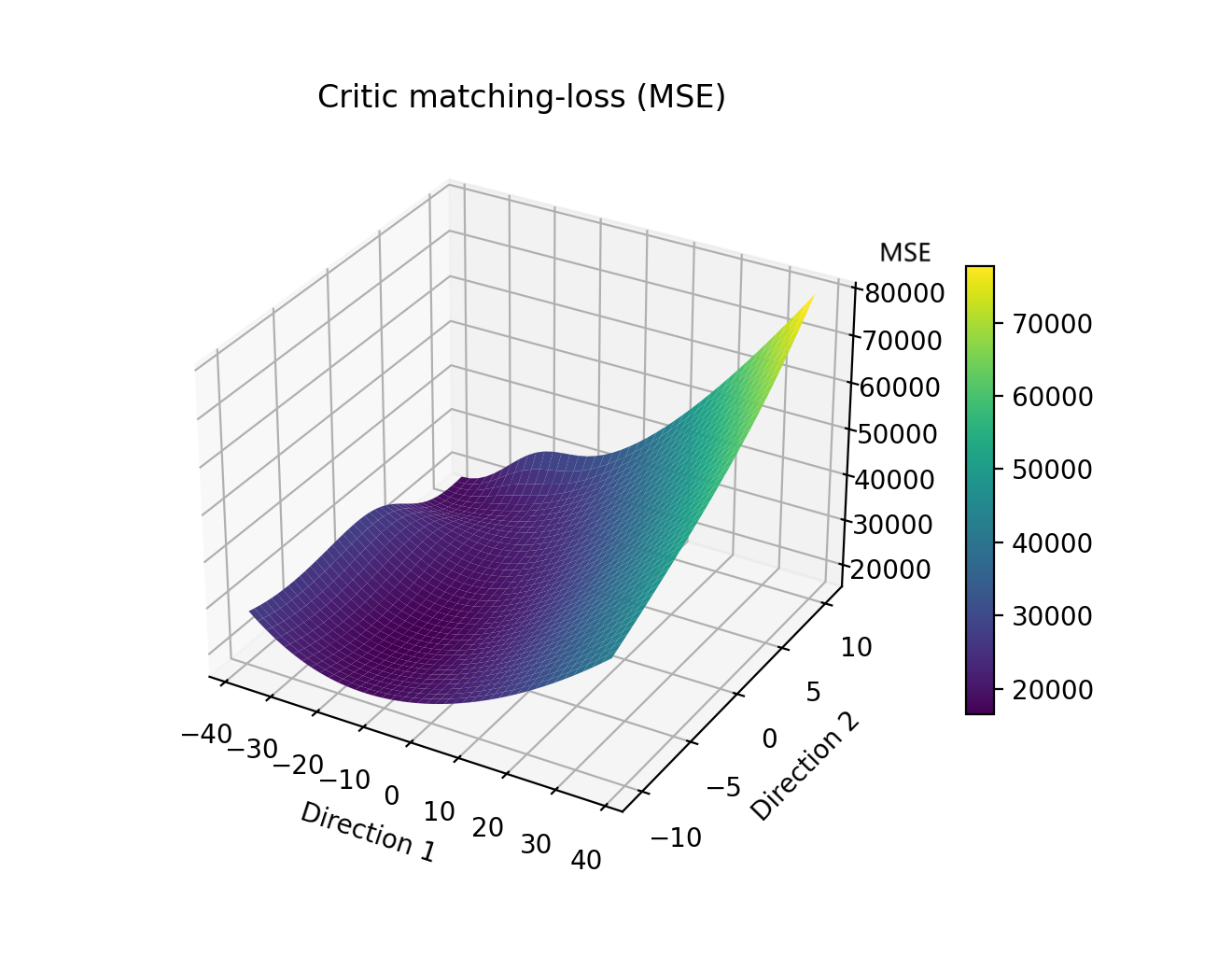}
         \caption{Critic Match Loss Landscape }
         \label{chapter4_fig:targetADHDPcritic}
     \end{subfigure}
     \begin{subfigure}[b]{0.49\textwidth}
         \centering
         \includegraphics[width=\textwidth]{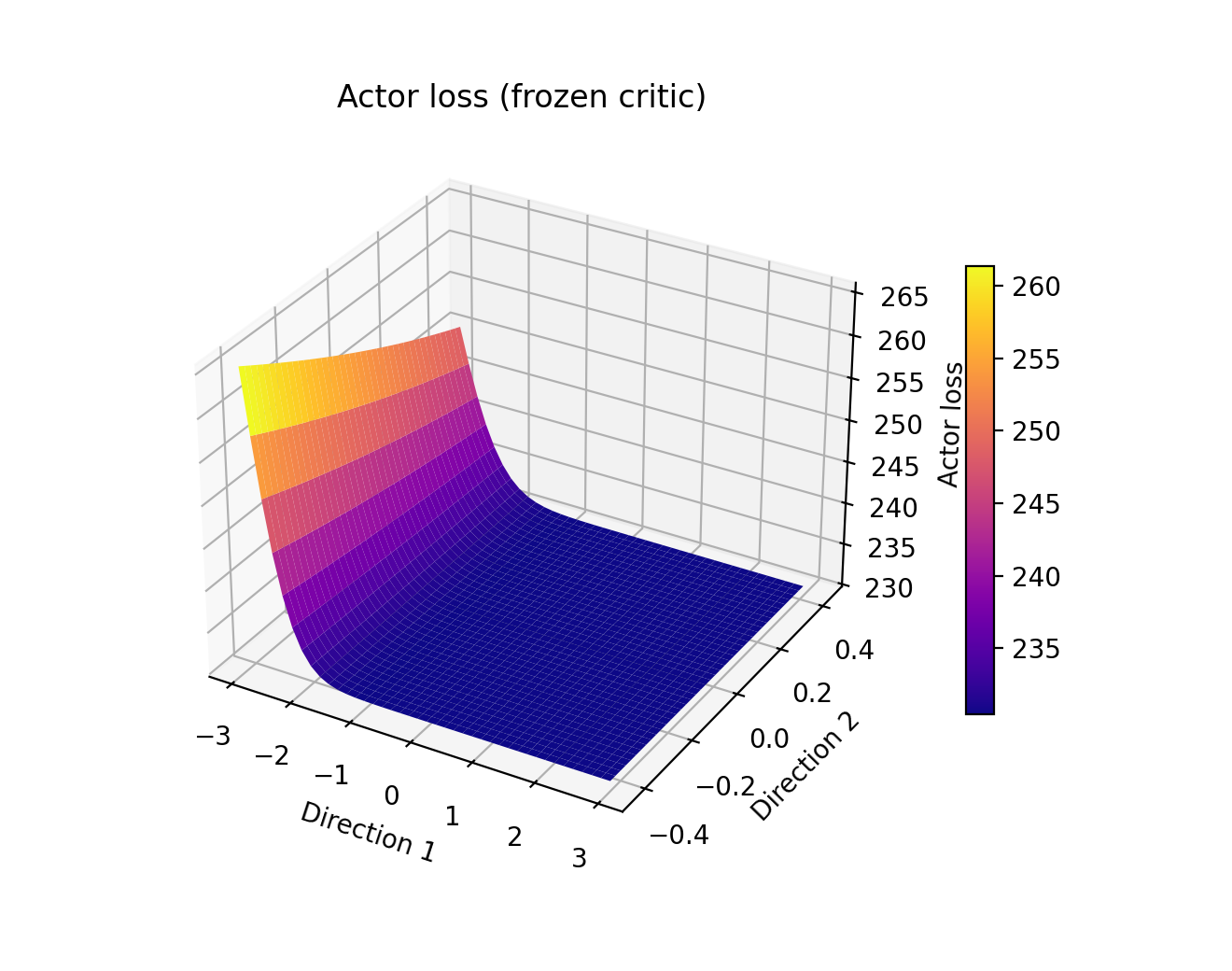}
         \caption{Actor Loss Landscape}
         \label{chapter4_fig:targetADHDPactorloss}
     \end{subfigure}
        \caption{Loss surface, ADHDP with target net}
        \label{chapter4_fig:targetADHDPlosssurface}
\end{figure}
In \autoref{chapter4_fig:targetADHDPcritic}, it shows that the critic match loss approximates a wide and shallow bowl, with an upward trend in the front right. In the upper right region of the critic match loss landscape, small changes in the critic weight can lead to large changes in the critic match error.  In \autoref{chapter4_fig:targetADHDPlosssurface}, it shows that the actor loss surface exhibits a typical wedge-shaped valley, with a steep slope along Direction-1 and a relatively flat slope along Direction-2. With a frozen critic, the actor loss exhibits a strong gradient along one direction but remains nearly flat along another direction. Under the action cap constraint, the weak gradient region is easily trapped at the saturation boundary, resulting in near-saturation of the control torque.

\begin{figure}[htbp]
     \centering
     \begin{subfigure}[b]{0.5\textwidth}
         \centering
         \includegraphics[width=\textwidth]{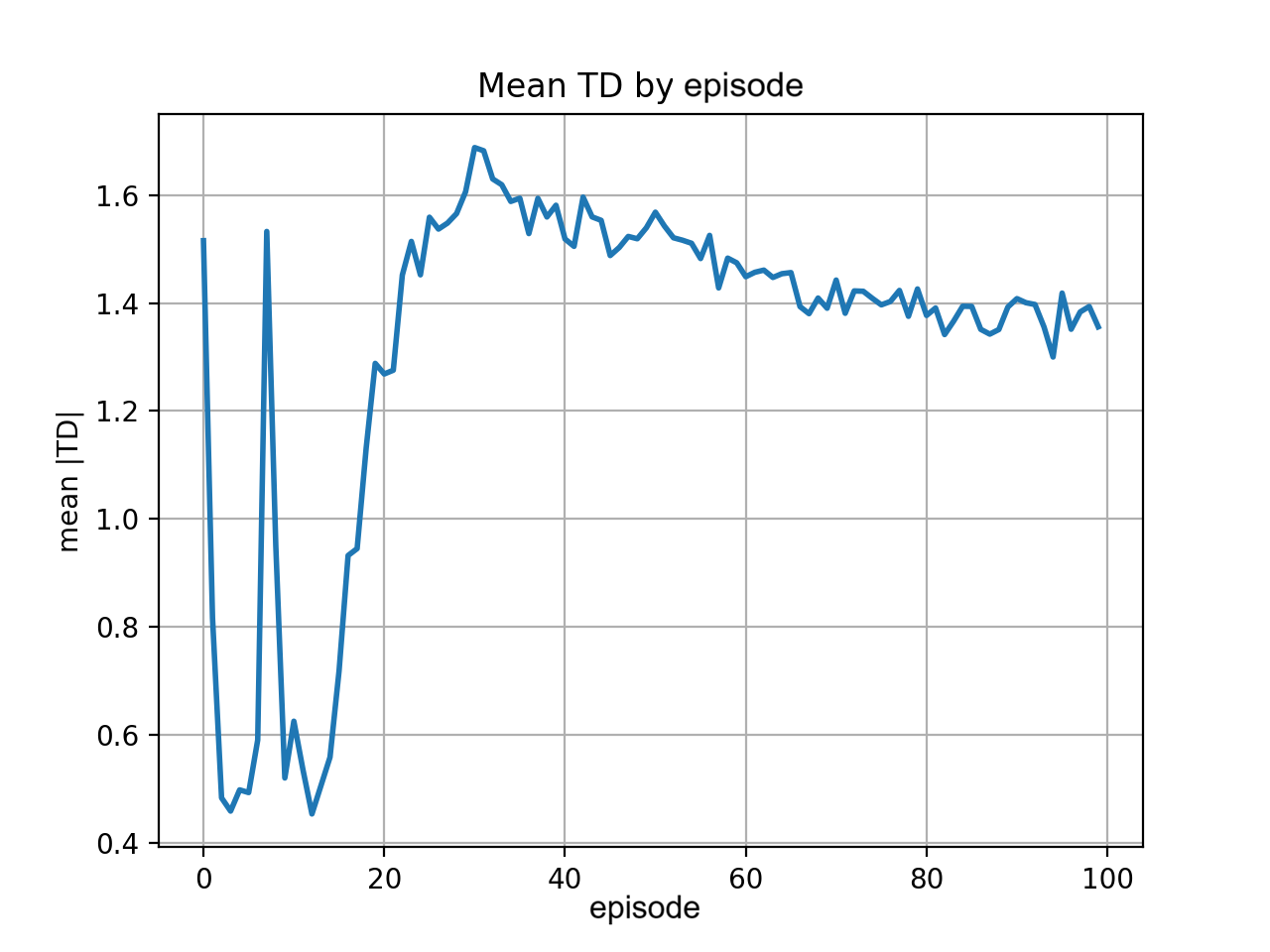}
         \caption{TD by episode}
         \label{chapter4_fig:targetADHDPTDbytrial}
     \end{subfigure}
     \hfill
     \begin{subfigure}[b]{0.65\textwidth}
         \centering
         \includegraphics[width=\textwidth]{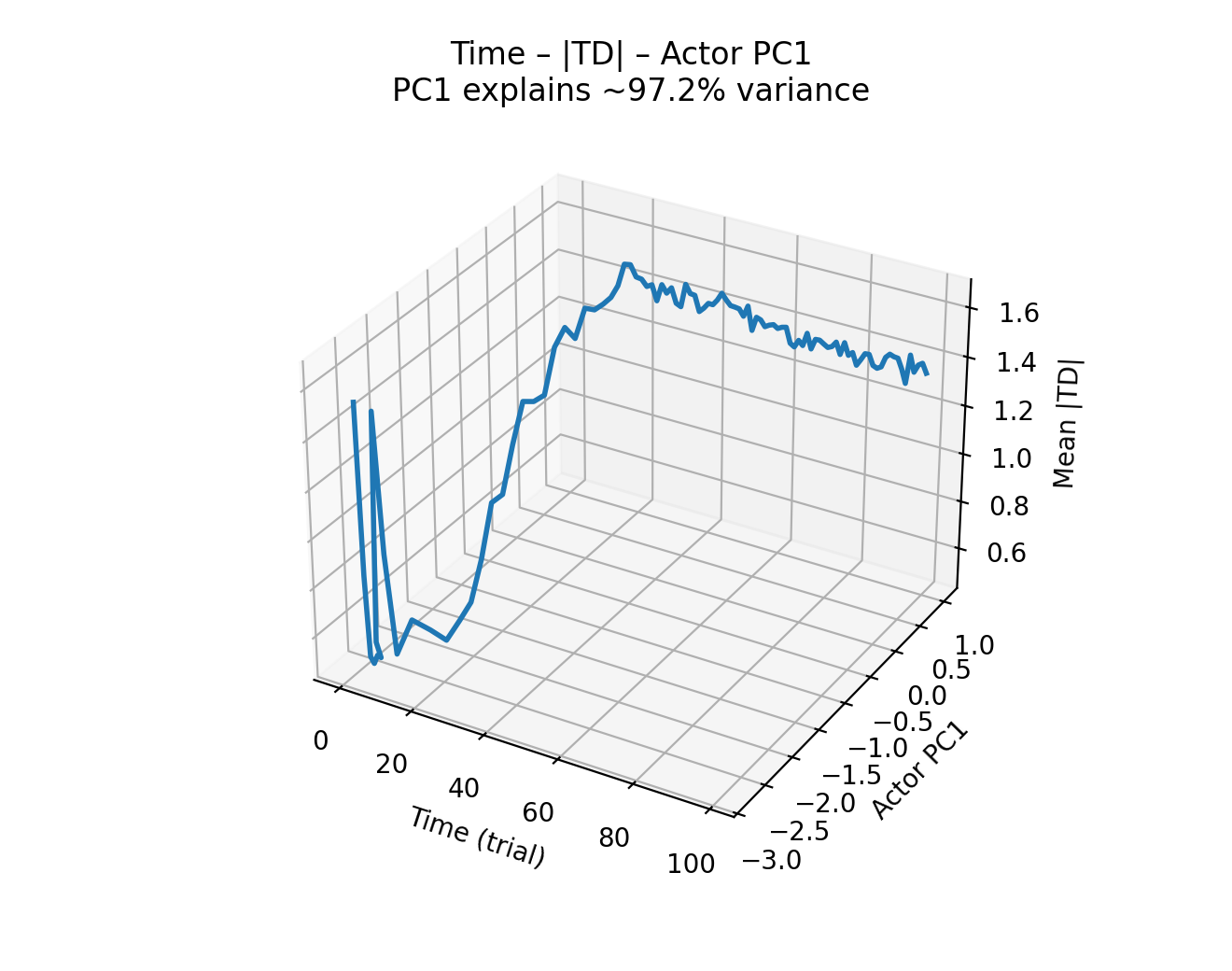}
         \caption{Actor weight with time and TD}
         \label{chapter4_fig:targetADHDPactweightTD}
     \end{subfigure}
     \hfill
     \begin{subfigure}[b]{0.5\textwidth}
         \centering
         \includegraphics[width=\textwidth]{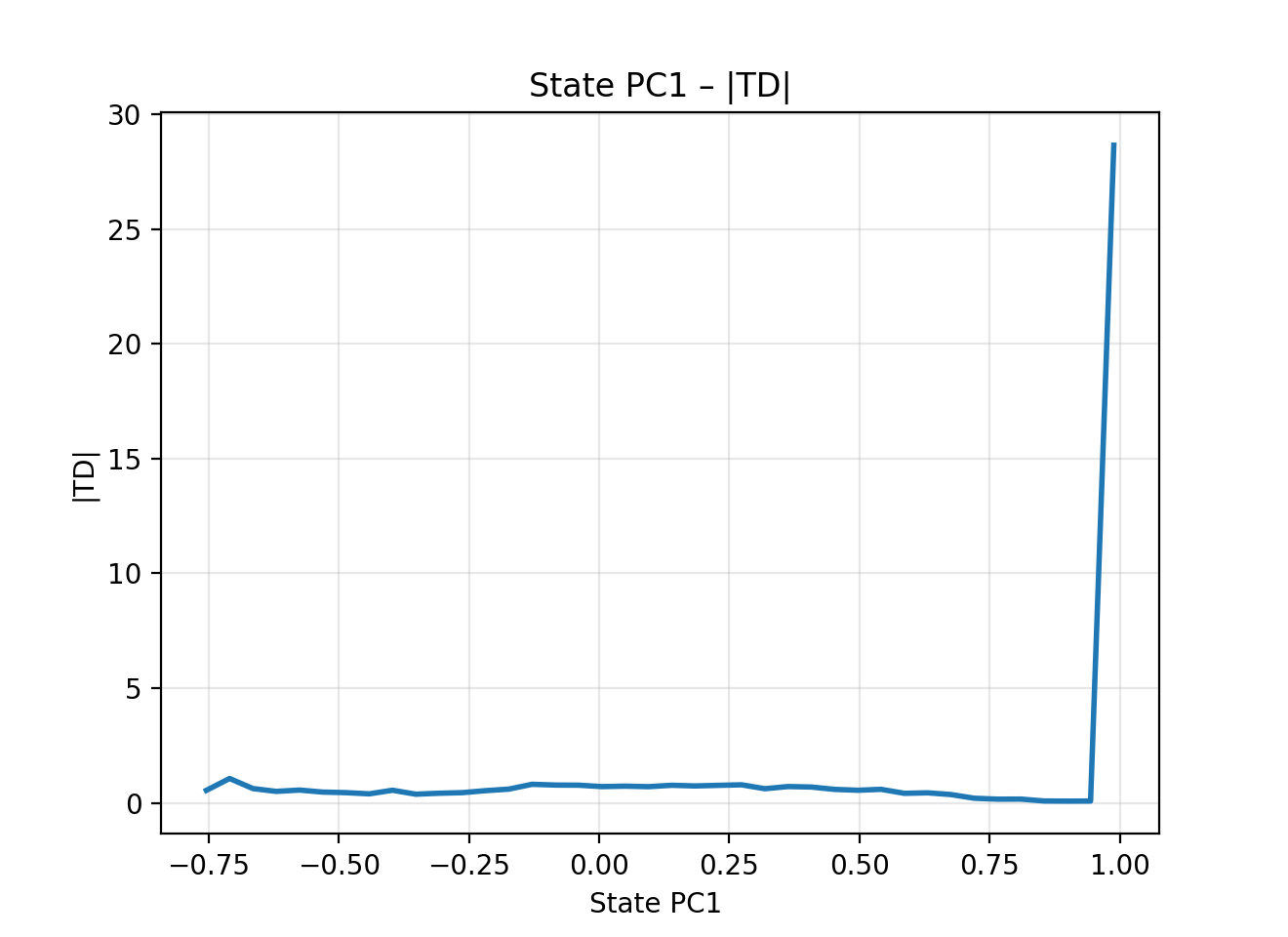}
         \caption{State with TD}
         \label{chapter4_fig:targetADHDPstateTD}
     \end{subfigure}
        \caption{TD trajectory, ADHDP with target net}
        \label{chapter4_fig:targetADHDPTD}
\end{figure}

In \autoref{chapter4_fig:targetADHDPTDbytrial}, it shows that the TD error increases rapidly in the early stages, then slowly decreases to a value that is still high in the middle stages. Compared to the basic ADHDP version, the TD error fluctuates within a narrower range throughout the entire process. This indicates that using the target net can mitigate jitter, but it cannot reduce abnormal TD error to a sufficiently small value to allow the states to converge to a stable value. In \autoref{chapter4_fig:targetADHDPstateTD}, it shows that the states have a huge spike at the end, indicating that the TD error becomes extremely large when the system approaches unstable regions of the state space. 

From the visualization framework, the reason for the control failure can be further explained. The critic match loss landscape in \autoref{chapter4_fig:targetADHDPcritic} becomes smoother in the loss compared with the basic ADHDP version, which shows that the target network stabilizes the critic fitting process. However, the surface still contains a long ridge region where the loss changes rapidly along one direction. When the actor updates its parameters under this critic, the policy gradient mainly follows the steep direction of the actor loss surface while remaining weak along the flat direction, as shown in  \autoref{chapter4_fig:targetADHDPlosssurface}. Under the action cap constraint, this imbalance easily drives the policy toward the saturation boundary. As a result, the control torques remain close to their limits, which can be observed in \autoref{chapter4_fig:targetADHDPstates}. Once the system enters these saturated control regions, the collected samples contain large TD errors, which further correspond to the spike observed in the state--TD plot in \autoref{chapter4_fig:targetADHDPstateTD}. Therefore, although the target network makes the TD curve smoother, the geometry of the critic and actor landscapes still leads the policy to saturated actions, and the closed-loop control eventually becomes unstable.

This consistency among the visualizations shows that the target network mainly slows down the update of TD targets. It helps to reduce short-term fluctuation but does not change the overall shape of the loss landscape or the distribution of the learning signals. From the visualization framework, it can be seen that the target network makes the training more stable, but it does not improve the convergence of the algorithm.

Although the target network improves the stability of the critic update, the previous analysis shows that it does not fundamentally resolve the unfavorable geometry of the loss landscapes. The critic match loss surface still contains elongated ridge regions, and the actor loss surface exhibits a strong imbalance between steep and flat directions. Under the action cap constraint, this imbalance easily drives the policy toward saturation, which eventually leads to unstable closed-loop behavior. To further improve the stability of the training process, additional stabilization techniques are introduced in the next version of ADHDP. Building upon the target-network framework, this version incorporates two training stabilizers, namely numeric cost scaling and the Huber critic loss. In addition, target policy smoothing is applied to reduce the sensitivity of the critic to small variations in the policy. The following analysis applies the same visualization framework to examine how these mechanisms influence the critic and actor training dynamics.

\subsection{ADHDP with training stabilizers and target policy smoothing}

Compared with the ADHDP with target net, this variant introduces several stabilizing and training techniques. It shares the same simulation set up as the ADHDP with target net, but it has two more stablization techniques and target policy smoothing. Huber loss for critic is used in this version.  Critic loss uses smooth $\ell_1$ Huber loss instead of standard MSE. Cost scaling is also used, which means that all immediate costs are scaled by factor $10.0$ to improve numerical stability without changing the optimal point.Target action smoothing is also added. Additive noise on the target action used in TD bootstrap. Smoothing noise standard deviation $\sigma_\text{noise, smooth}$ is set to 5\% of the torque limit used in the controller..

\begin{figure}[htbp]
     \centering
     \begin{subfigure}[b]{0.55\textwidth}
         \centering
         \includegraphics[width=\textwidth]{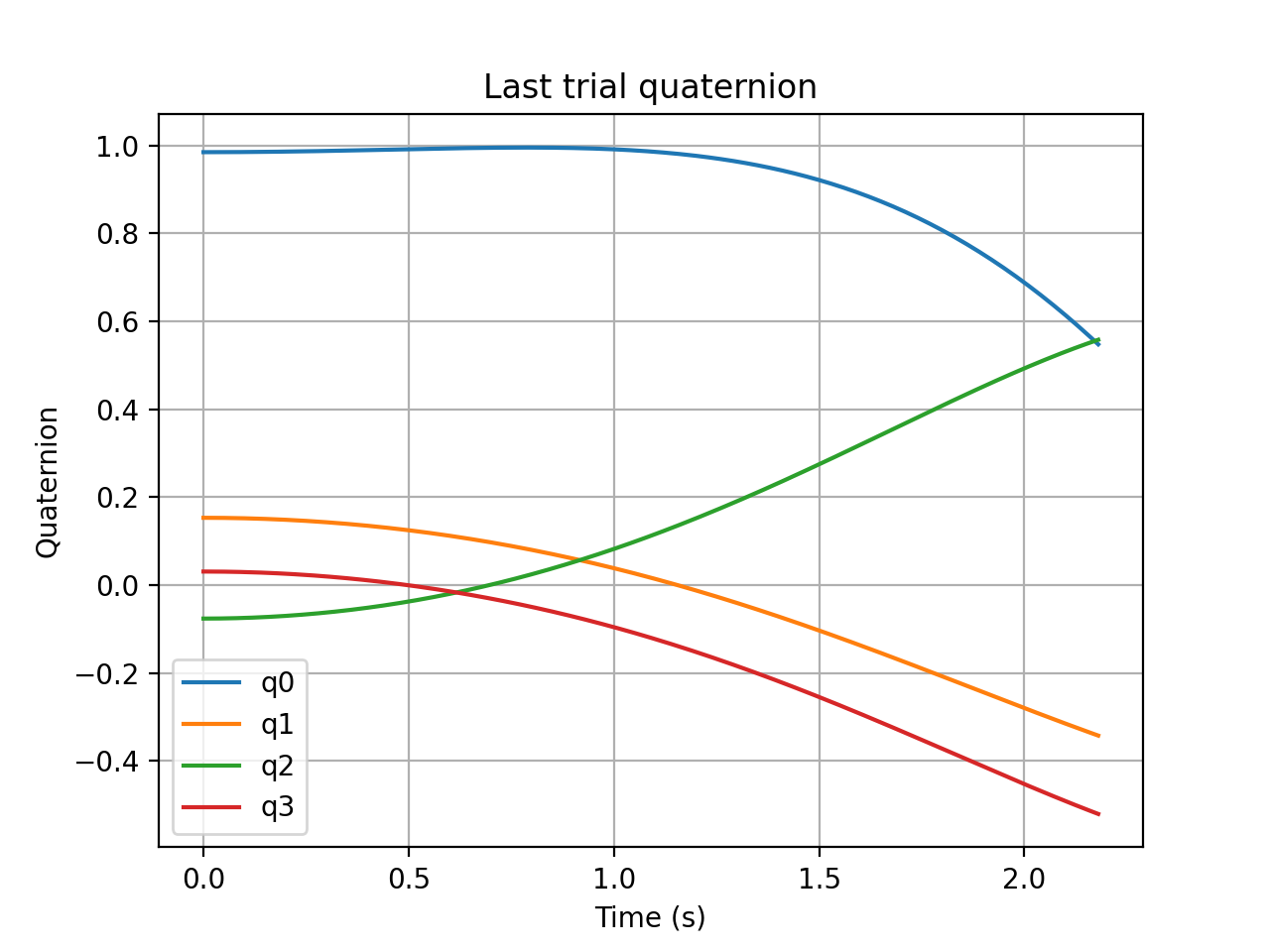}
         \caption{Quaternions}
         \label{chapter4_fig:advancedADHDPquat}
     \end{subfigure}
     \hfill
     \begin{subfigure}[b]{0.55\textwidth}
         \centering
         \includegraphics[width=\textwidth]{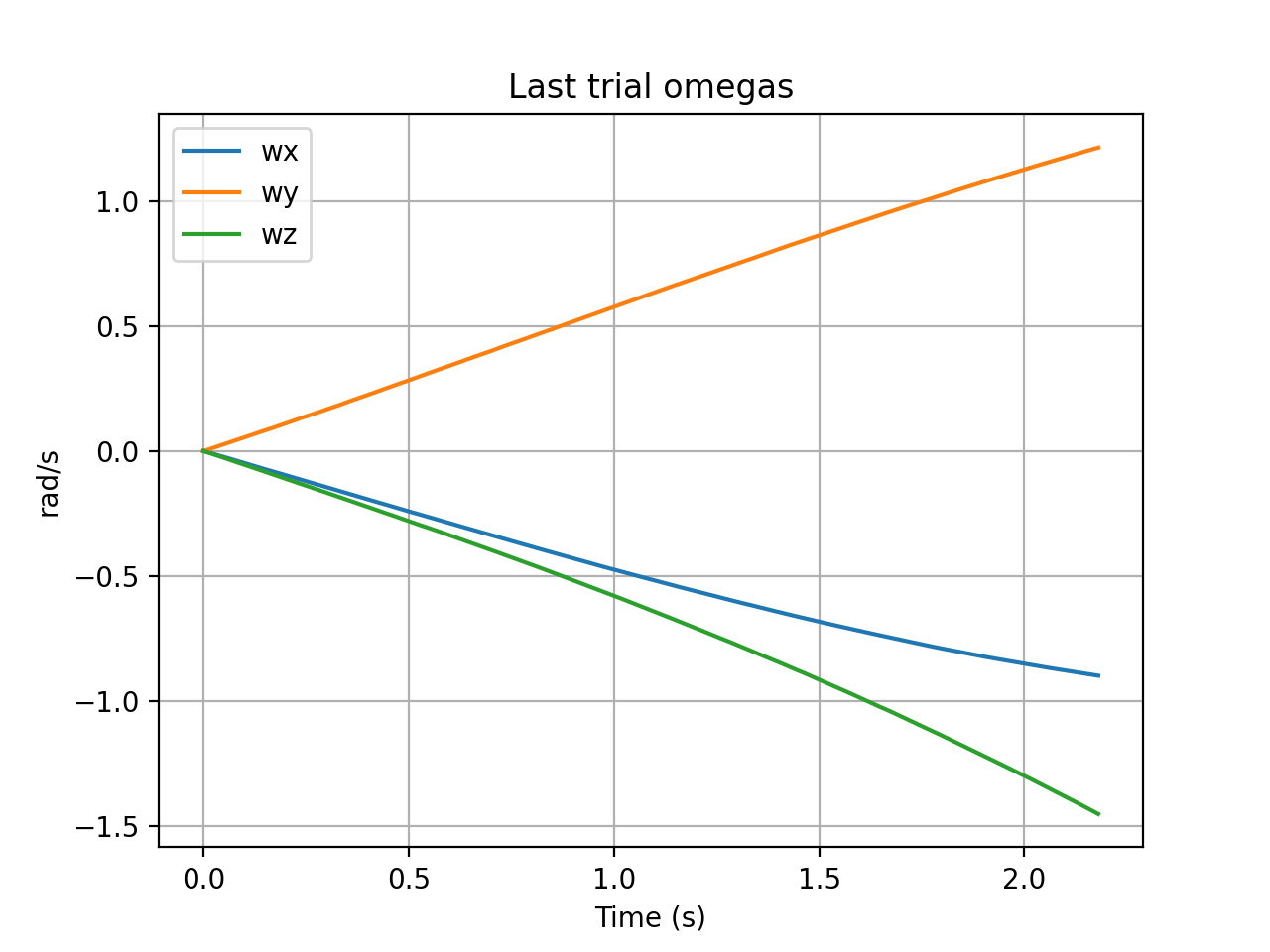}
         \caption{Angular velocity}
         \label{chapter4_fig:advancedADHDPomega}
     \end{subfigure}
     \hfill
     \begin{subfigure}[b]{0.55\textwidth}
         \centering
         \includegraphics[width=\textwidth]{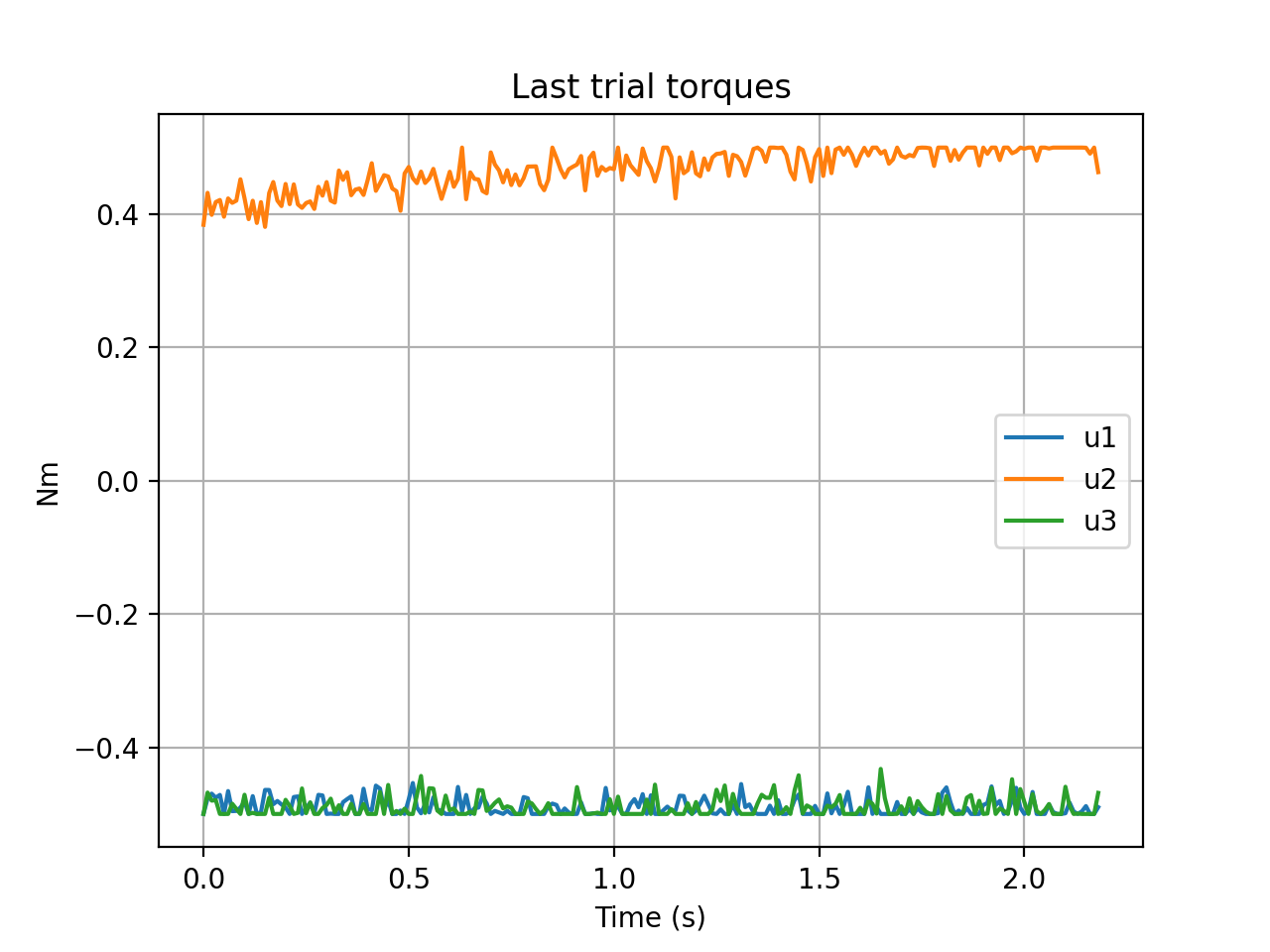}
         \caption{Control Torque}
         \label{chapter4_fig:advancedADHDPtorque}
     \end{subfigure}
        \caption{ ADHDP with training stabilizers (Huber critic loss and numeric cost scaling) and target policy smoothing}
        \label{chapter4_fig:advancedADHDPstates}
\end{figure}
With the version of the ADHDP algorithm training stabilizers and target policy smoothing, \autoref{chapter4_fig:advancedADHDPstates} shows the quaternions, angular velocities, and control torque of the system during the last episode of training. It shows the failed control result. With the figures below, the reasons for failure will be given.

\begin{figure}[htbp]
     \centering
     \begin{subfigure}[b]{0.49\textwidth}
         \centering
         \includegraphics[width=\textwidth]{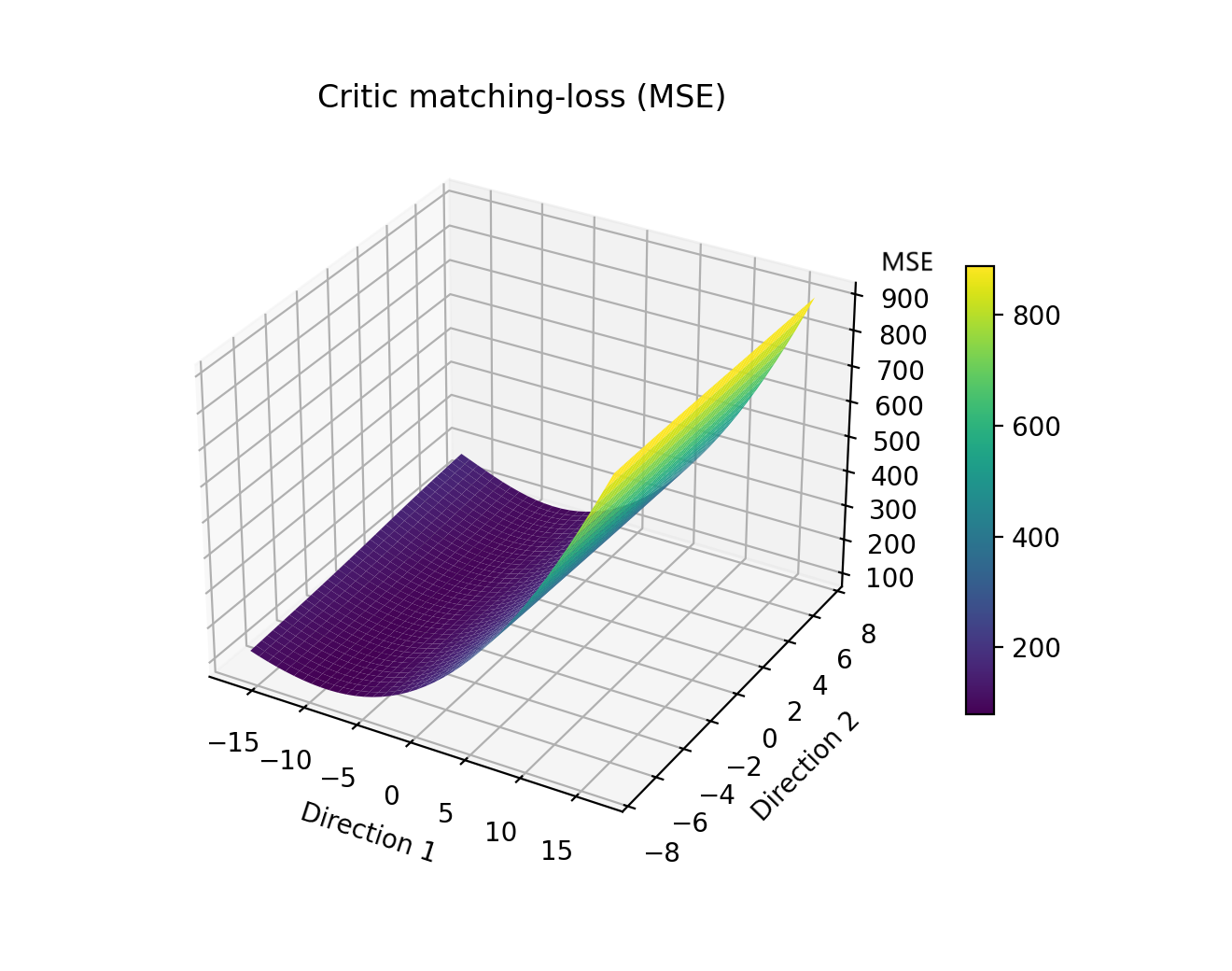}
         \caption{Critic Match Loss Landscape }
         \label{chapter4_fig:advancedADHDPcritic}
     \end{subfigure}
     \begin{subfigure}[b]{0.49\textwidth}
         \centering
         \includegraphics[width=\textwidth]{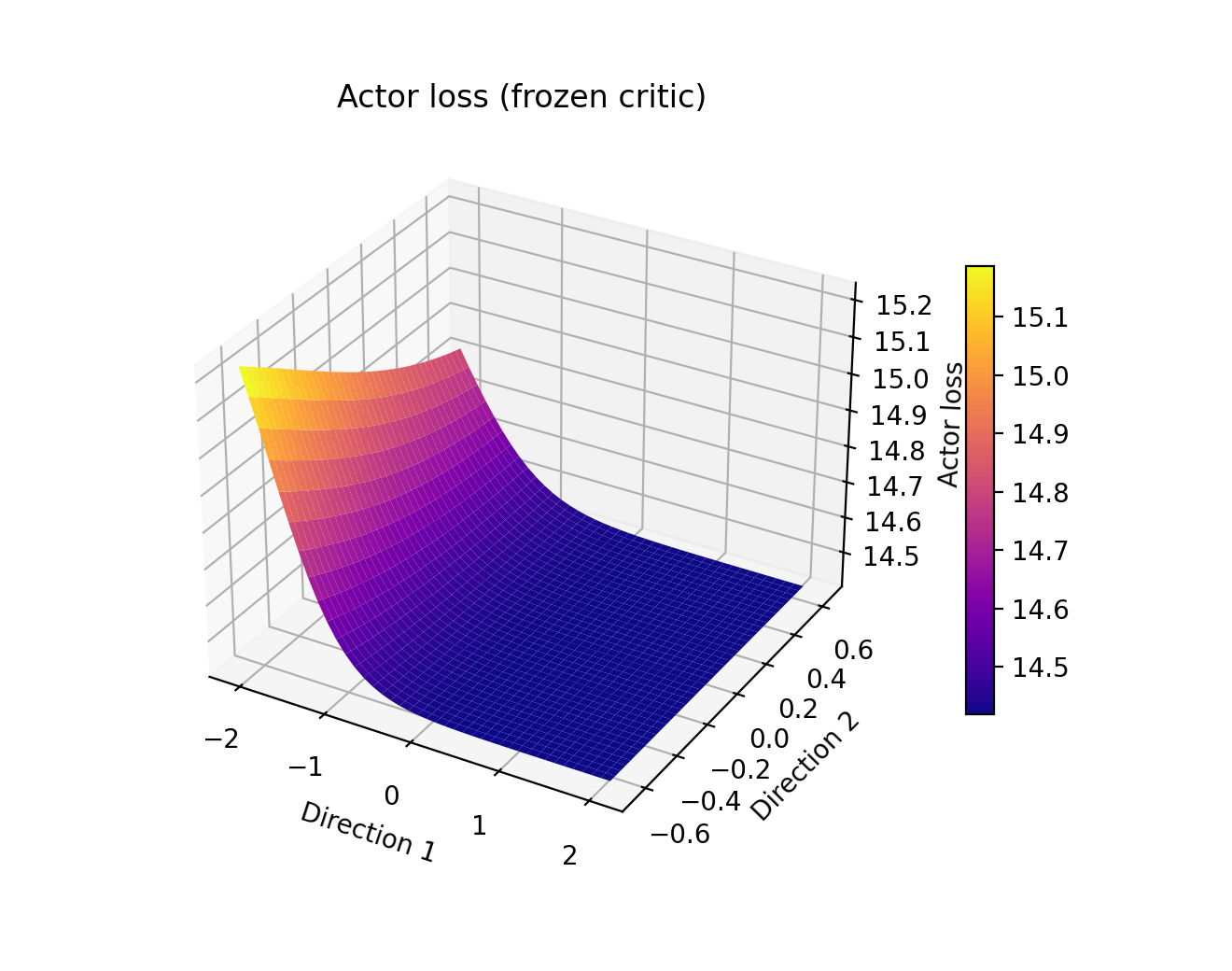}
         \caption{Actor Loss Landscape}
         \label{chapter4_fig:advancedADHDPactorloss}
     \end{subfigure}
        \caption{Loss surface, ADHDP with training stabilizers (Huber critic loss and numeric cost scaling) and target policy smoothing}
        \label{chapter4_fig:advancedADHDPlosssurface}
\end{figure}
In \autoref{chapter4_fig:advancedADHDPcritic}, it shows that the critic match loss landscape is more rounded and blunt compared to the ADHDP with the target net version, with no obvious long ridges. This indicates that the introduction of the Huber loss formulation and cost scaling truncates anomalous TD gradients. Together with target policy smoothing, the gradients in the critic loss landscape become more gentle, making it easier to learn. In \autoref{chapter4_fig:advancedADHDPlosssurface}, it shows that the actor loss landscape still exhibits a wedge shape, but with a smoother valley. This is consistent with the smoother and rounder nature of the critic match loss landscape.

\begin{figure}[htbp]
     \centering
     \begin{subfigure}[b]{0.5\textwidth}
         \centering
         \includegraphics[width=\textwidth]{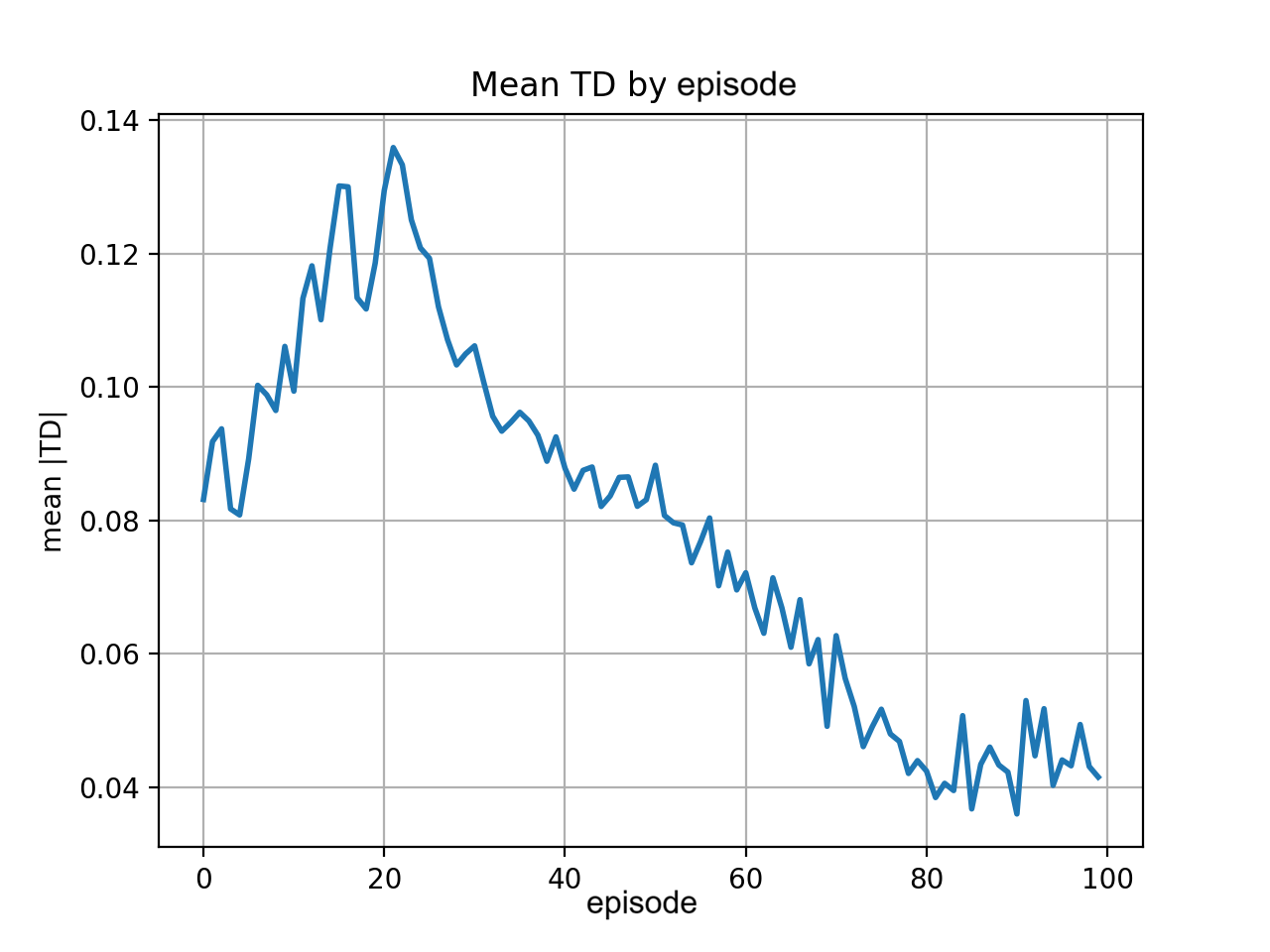}
         \caption{TD by episode}
         \label{chapter4_fig:advancedADHDPTDbytrial}
     \end{subfigure}
     \hfill
     \begin{subfigure}[b]{0.65\textwidth}
         \centering
         \includegraphics[width=\textwidth]{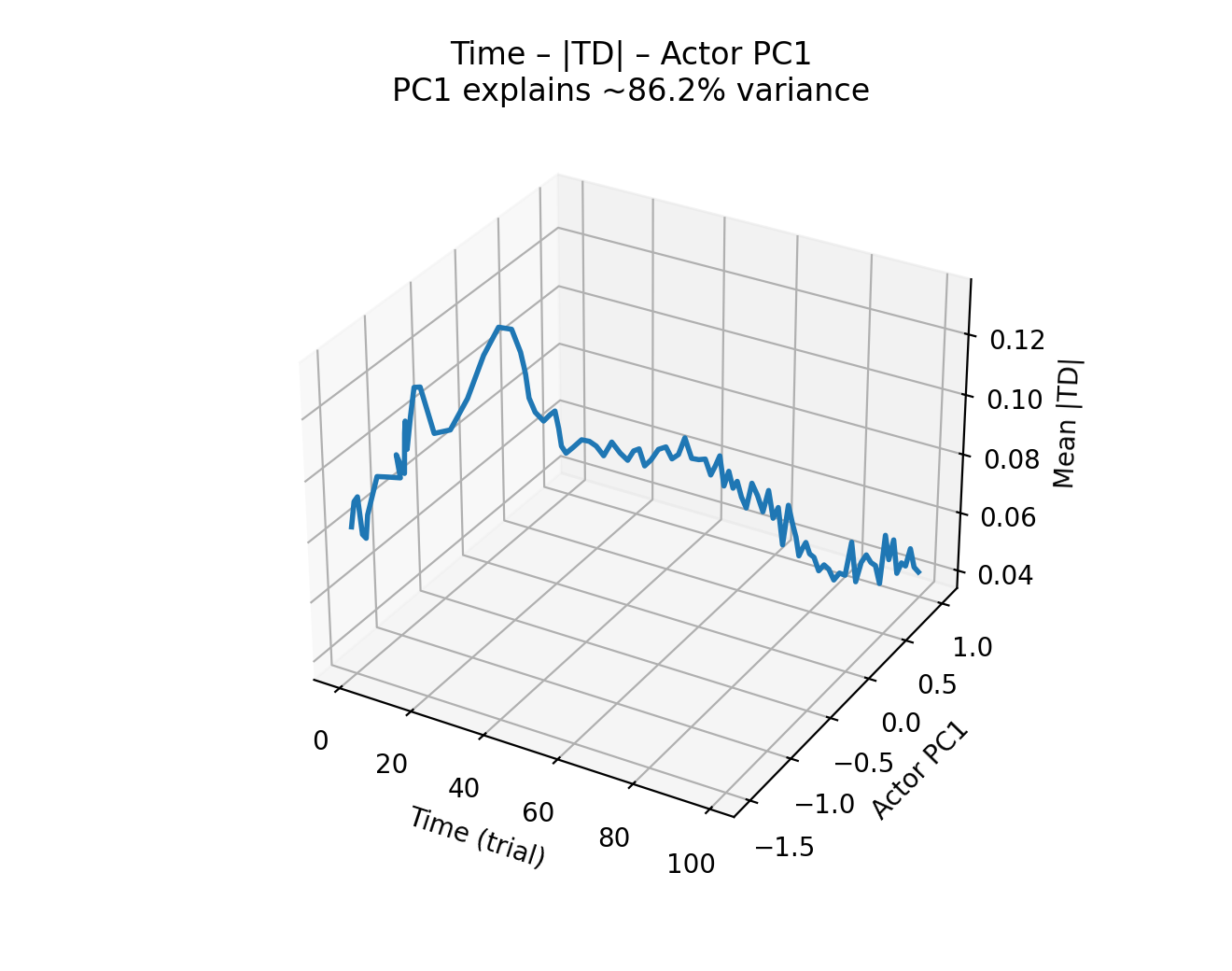}
         \caption{Actor weight with time and TD}
         \label{chapter4_fig:advancedADHDPactweightTD}
     \end{subfigure}
     \hfill
     \begin{subfigure}[b]{0.5\textwidth}
         \centering
         \includegraphics[width=\textwidth]{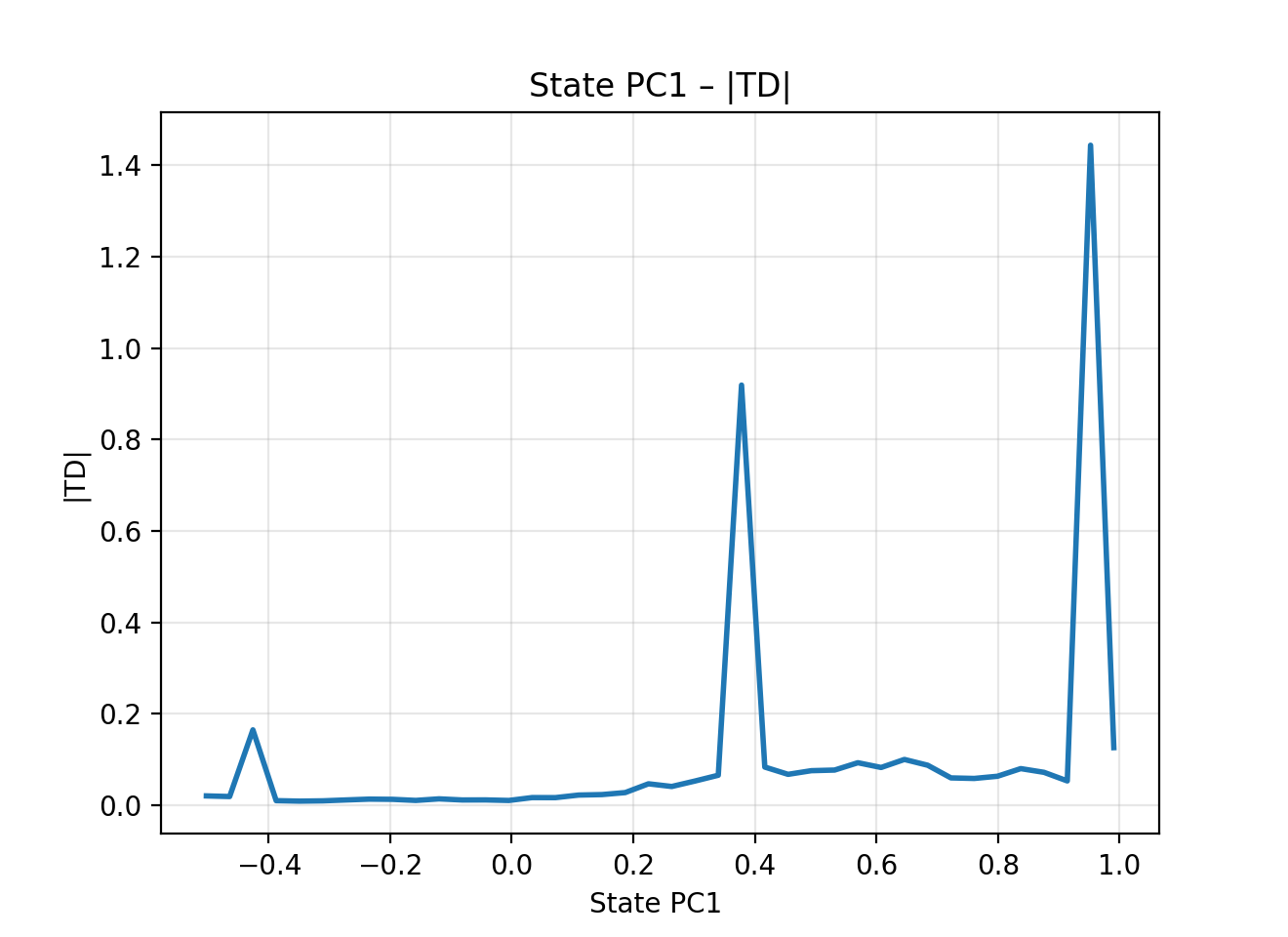}
         \caption{State with TD}
         \label{chapter4_fig:advancedADHDPstateTD}
     \end{subfigure}
        \caption{TD trajectory,  ADHDP with training stabilizers (Huber critic loss and numeric cost scaling) and target policy smoothing}
        \label{chapter4_fig:advancedADHDPTD}
\end{figure}

In \autoref{chapter4_fig:advancedADHDPTDbytrial}, it shows that the TD variation is smaller than that of the basic ADHDP version and the ADHDP with target net, gradually converging to smaller values. In \autoref{chapter4_fig:advancedADHDPactweightTD}, it shows that the first direction of actor weight exhibits a monotonically drifting trend while TD decreases during training. The PCA variance decreases from 97\%, which is the variance in the target ADHDP version, to 86\%, indicating that the weight evolution is still dominated by one main direction. In \autoref{chapter4_fig:advancedADHDPstateTD}, medium-sized peaks and a flatter surface are shown compared to the basic ADHDP version and the ADHDP with target net version, indicating that extreme endpoints are suppressed by the Huber loss and cost scaling.

However, the visualization framework also explains why the control still fails. Although the critic loss landscape in \autoref{chapter4_fig:advancedADHDPcritic} becomes smoother and more rounded after introducing Huber loss, cost scaling, and target policy smoothing, the global geometry of the landscape still contains a dominant slope along Direction-1. As a result, the actor loss surface in \autoref{chapter4_fig:advancedADHDPactorloss} still exhibits a wedge-shaped valley, although the valley becomes smoother. This means that the policy gradient remains much stronger in one direction than in the other. During training, the actor parameters therefore drift mainly along this dominant direction, which is consistent with the trajectory shown in \autoref{chapter4_fig:advancedADHDPactweightTD}. Although the TD magnitude decreases during training, the policy update is still biased toward this direction and cannot sufficiently explore other directions of the policy space. Under the torque saturation constraint, this biased update easily pushes the policy toward the saturation boundary. Consequently, the control torques remain close to their limits, which can be observed in \autoref{chapter4_fig:advancedADHDPstates}. Once the control inputs approach saturation, the spacecraft states cannot be corrected effectively and the attitude error gradually accumulates, eventually leading to divergence of the closed-loop control. This also explains why the TD curves appear smoother while the system trajectories in \autoref{chapter4_fig:advancedADHDPstates} still fail to converge.

In conclusion, the Huber loss truncates the gradients of abnormal TD, significantly reducing the range of TD error and the peaks in state-TD. The cost scaling technique avoids excessive fluctuations in the loss landscape, making it easier to train the actor and critic weights. Target policy smoothing makes the critic loss landscape more rounded and the actor loss landscape more flat, allowing the TD error to decrease more smoothly during training. The combined effect of multiple stabilizers improves training metrics. However, these improvements do not fundamentally change the anisotropic geometry of the actor loss landscape, which still drives the policy toward saturated actions.

Although the combination of training stabilizers and target policy smoothing improves the smoothness of the loss landscapes and reduces the TD error magnitude, the closed-loop control still fails to converge. Since several stabilization mechanisms are introduced simultaneously in the previous version, it is important to further examine the specific role of target policy smoothing in the training dynamics. Therefore, in the next experiment, the target policy smoothing mechanism is removed while retaining the two training stabilizers, namely the Huber critic loss and cost scaling. Using the same visualization framework, the resulting changes in the critic loss landscape, actor optimization trajectory, and TD distribution are analyzed.

\subsection{ADHDP with training stabilizers}
In this version, the target policy smoothing is removed compared with the version of ADHDP with training stabilizers and target policy smoothing. The control results and visualization results are shown in the following figures.
\begin{figure}[htbp]
     \centering
     \begin{subfigure}[b]{0.55\textwidth}
         \centering
         \includegraphics[width=\textwidth]{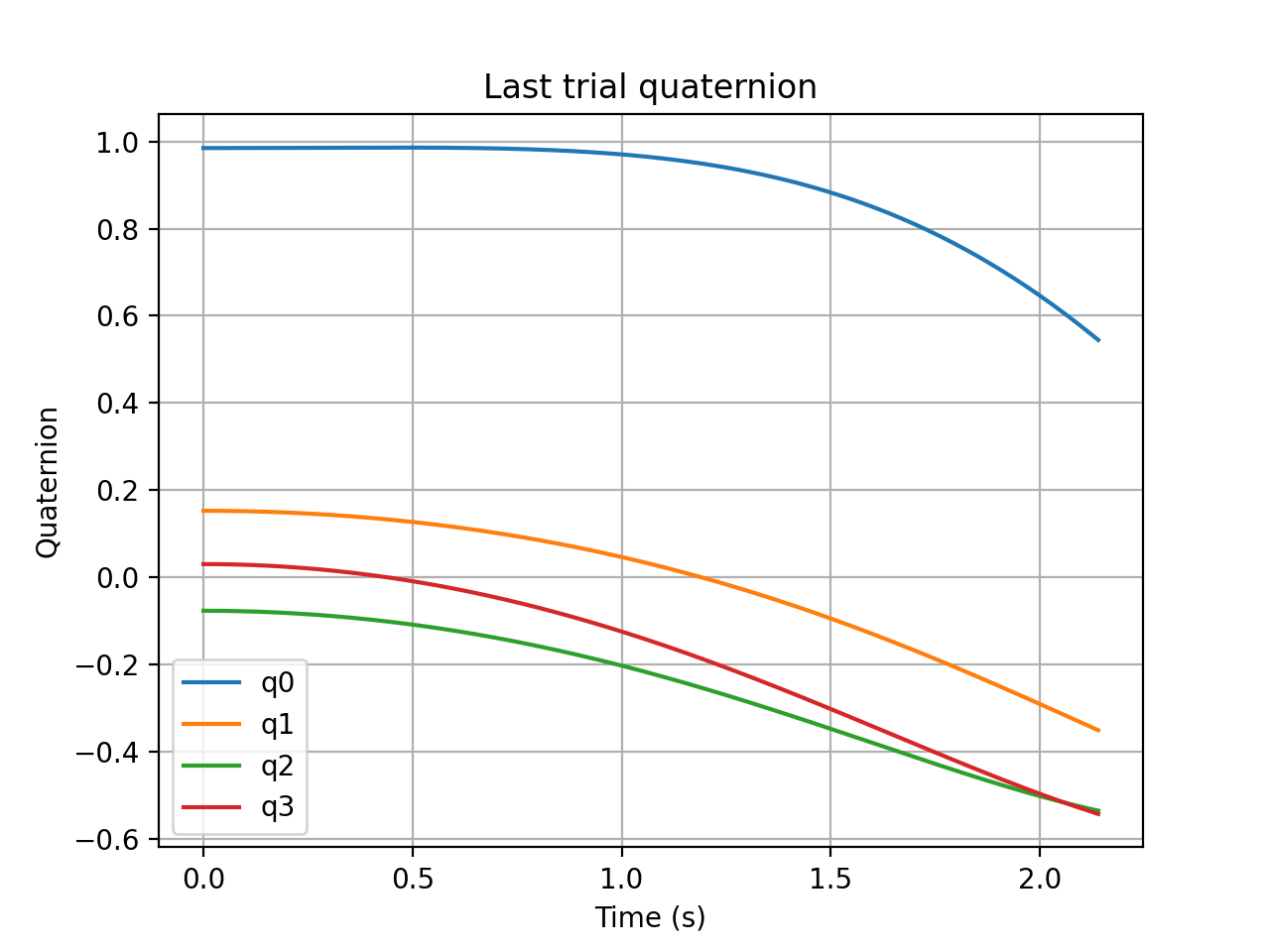}
         \caption{Quaternions}
         \label{chapter4_fig:target_nosmoothingADHDPquat}
     \end{subfigure}
     \hfill
     \begin{subfigure}[b]{0.55\textwidth}
         \centering
         \includegraphics[width=\textwidth]{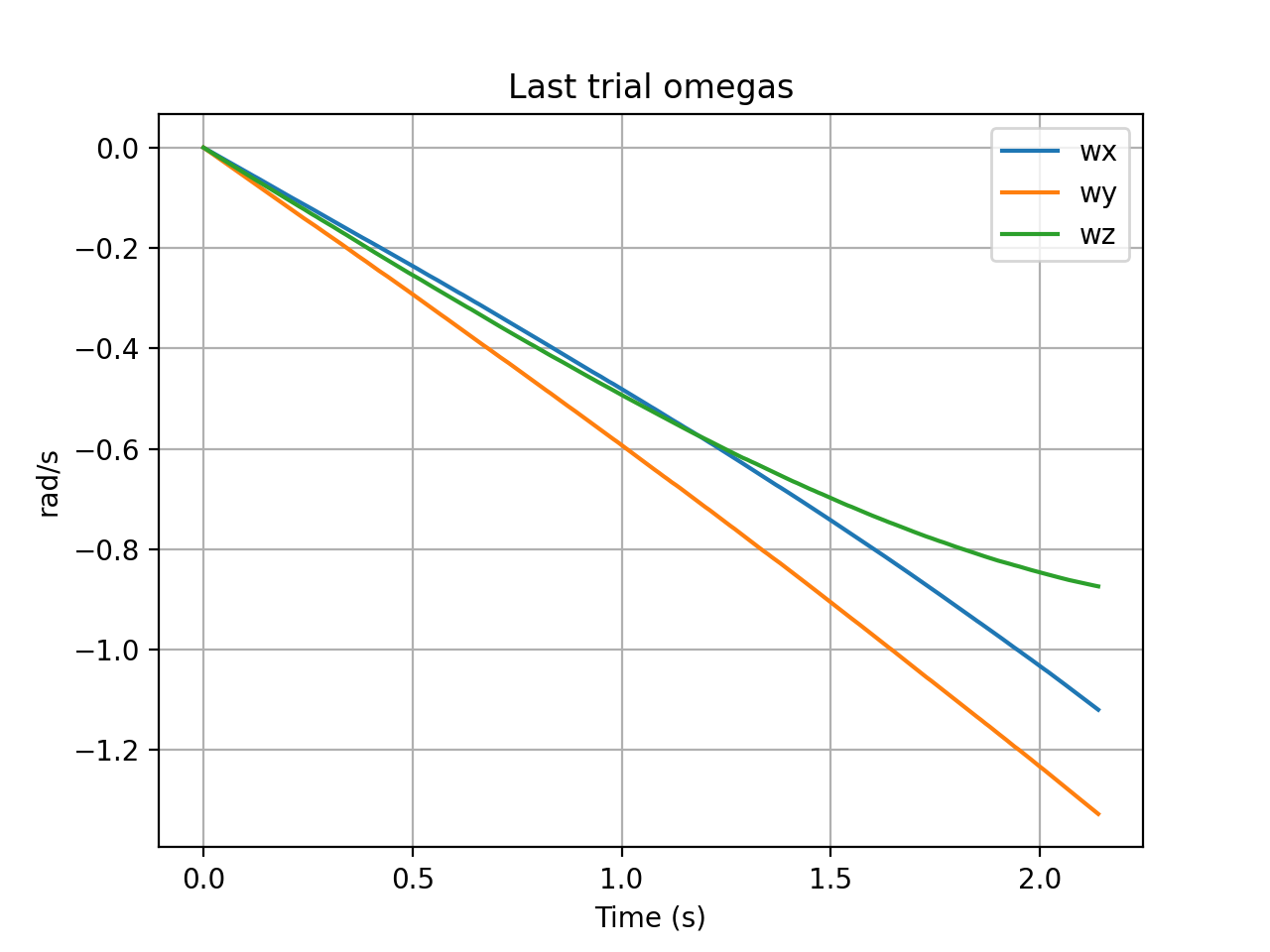}
         \caption{Angular velocity}
         \label{chapter4_fig:target_nosmoothingADHDPomega}
     \end{subfigure}
     \hfill
     \begin{subfigure}[b]{0.55\textwidth}
         \centering
         \includegraphics[width=\textwidth]{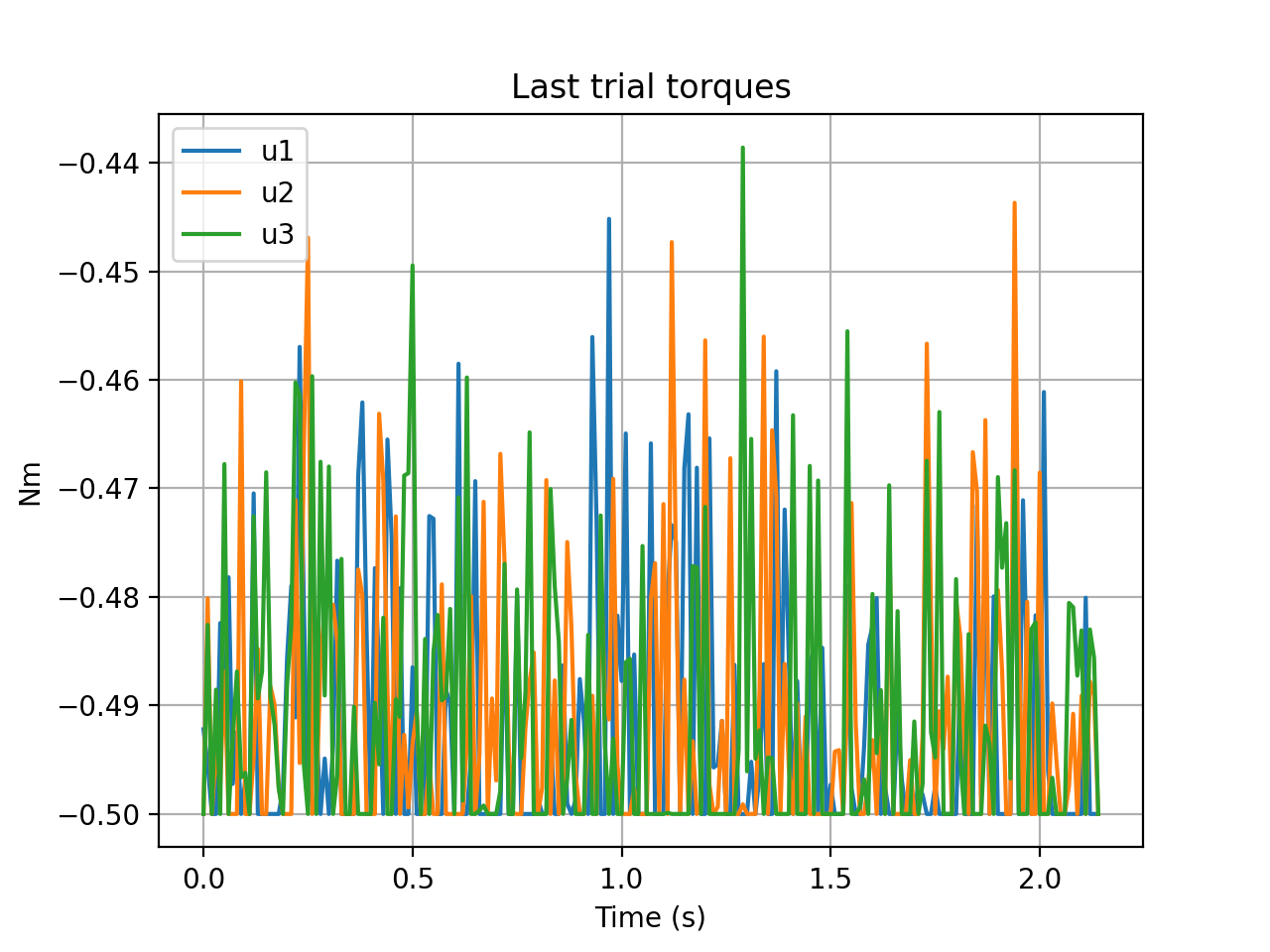}
         \caption{Control Torque}
         \label{chapter4_fig:target_nosmoothingADHDPtorque}
     \end{subfigure}
        \caption{State trajectory, ADHDP with training stabilizers(Huber critic loss and numeric cost scaling), no target policy smoothing}
        \label{chapter4_fig:target_nosmoothingADHDPstates}
\end{figure}

With the version of the ADHDP algorithm training stabilizers, \autoref{chapter4_fig:target_nosmoothingADHDPstates} shows the quaternions, angular velocities, and control torque of the system during the last episode of training. It shows the failed control result. With the figures below, the reasons for failure will be given.

\begin{figure}[htbp]
     \centering
     \begin{subfigure}[b]{0.49\textwidth}
         \centering
         \includegraphics[width=\textwidth]{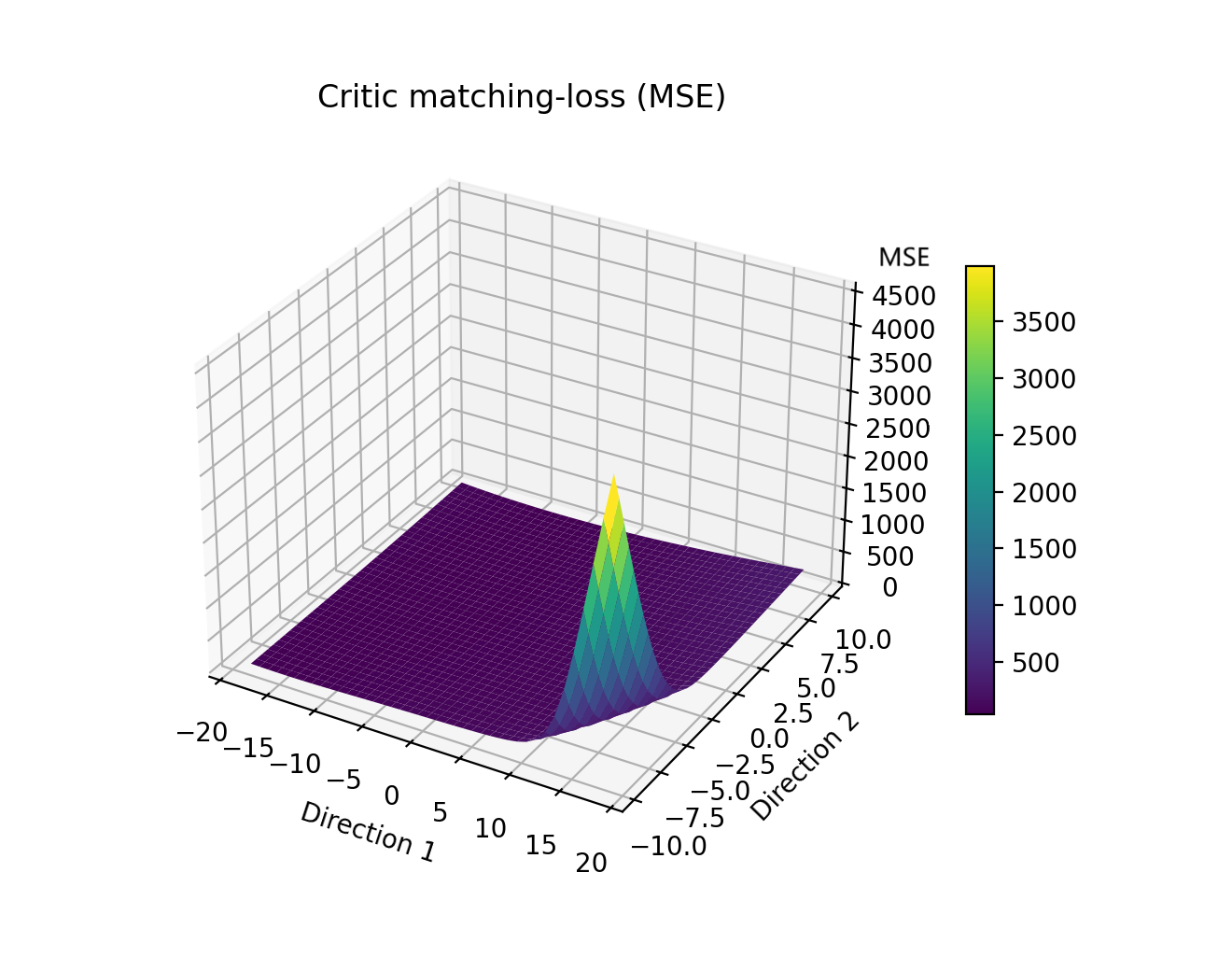}
         \caption{Critic Match Loss Landscape }
         \label{chapter4_fig:target_nosmoothingADHDPcritic}
     \end{subfigure}
     \begin{subfigure}[b]{0.49\textwidth}
         \centering
         \includegraphics[width=\textwidth]{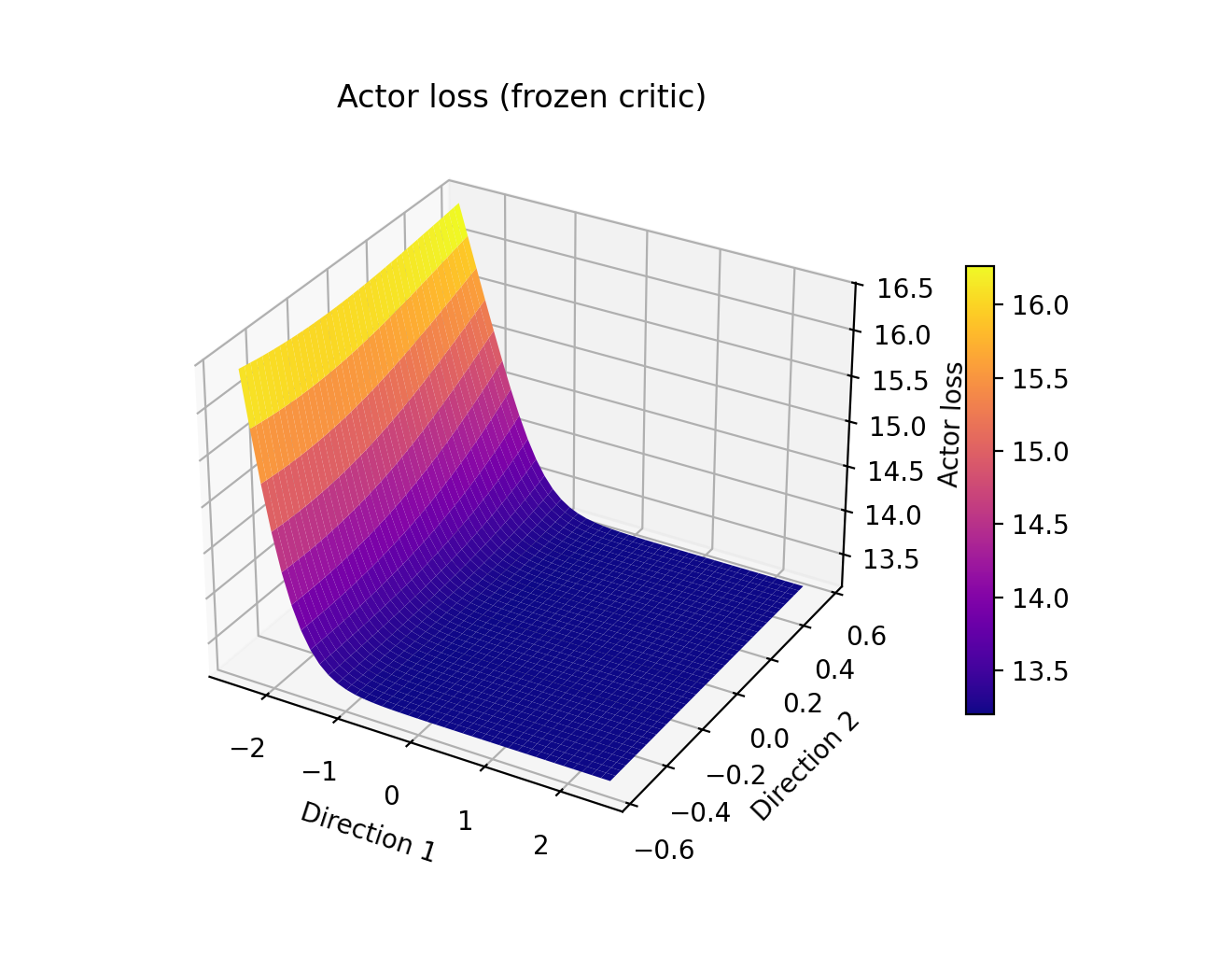}
         \caption{Actor Loss Landscape}
         \label{chapter4_fig:target_nosmoothingADHDPactorloss}
     \end{subfigure}
        \caption{Loss surface, ADHDP with training stabilizers(Huber critic loss and numeric cost scaling), no target policy smoothing}
        \label{chapter4_fig:target_nosmoothingADHDPlosssurface}
\end{figure}

In \autoref{chapter4_fig:target_nosmoothingADHDPcritic}, it shows that the critic match loss landscape exhibits a very sharp and narrow peak, much sharper than the ADHDP version with training stabilizer and target smoothing. This is due to the sharp terrain induced by the deterministic target without target policy smoothing.
The actor loss landscape still exhibits a wedge shape.

\begin{figure}[htbp]
     \centering
     \begin{subfigure}[b]{0.5\textwidth}
         \centering
         \includegraphics[width=\textwidth]{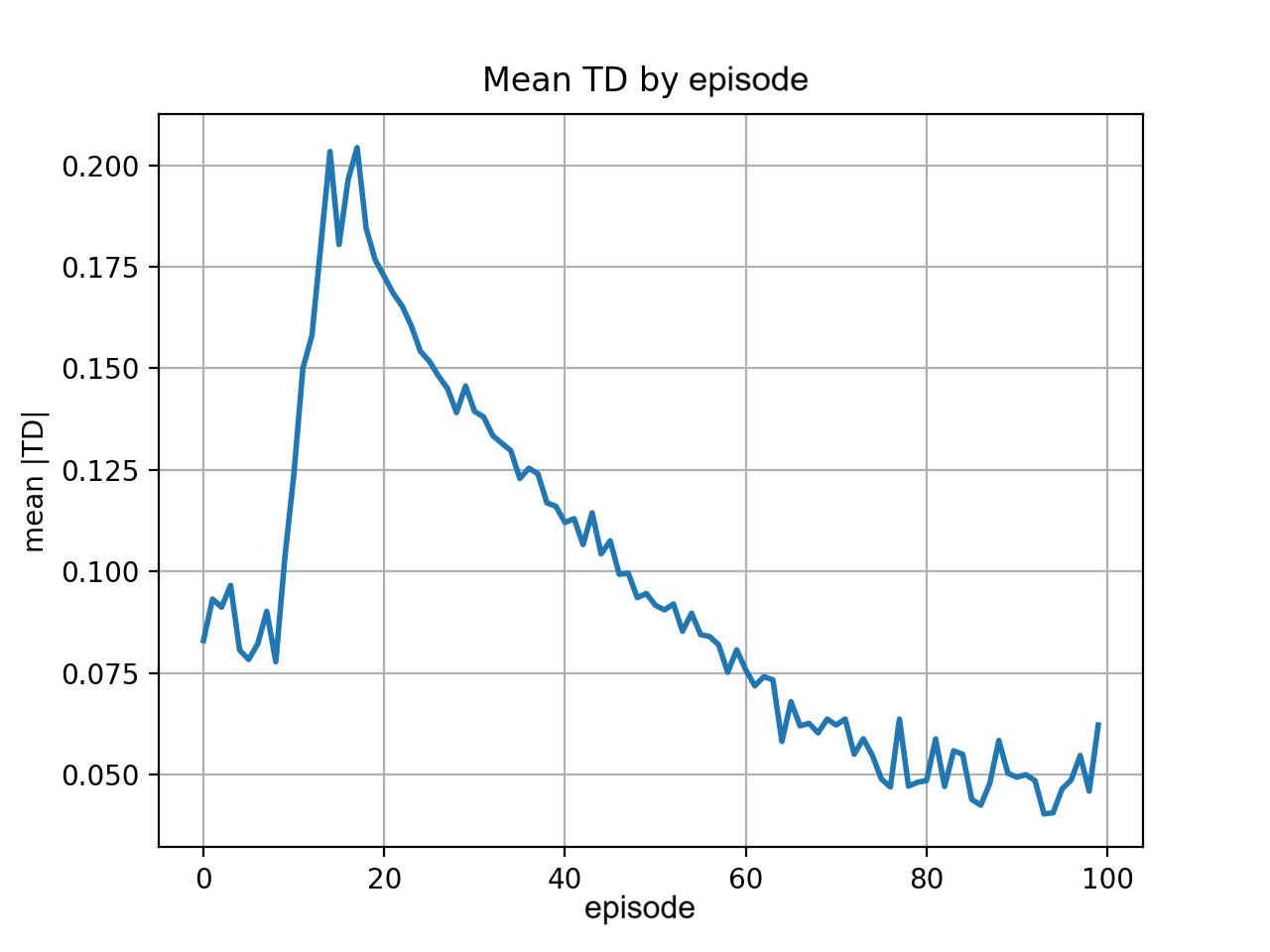}
         \caption{TD by episode}
         \label{chapter4_fig:target_nosmoothingADHDPTDbytrial}
     \end{subfigure}
     \hfill
     \begin{subfigure}[b]{0.65\textwidth}
         \centering
         \includegraphics[width=\textwidth]{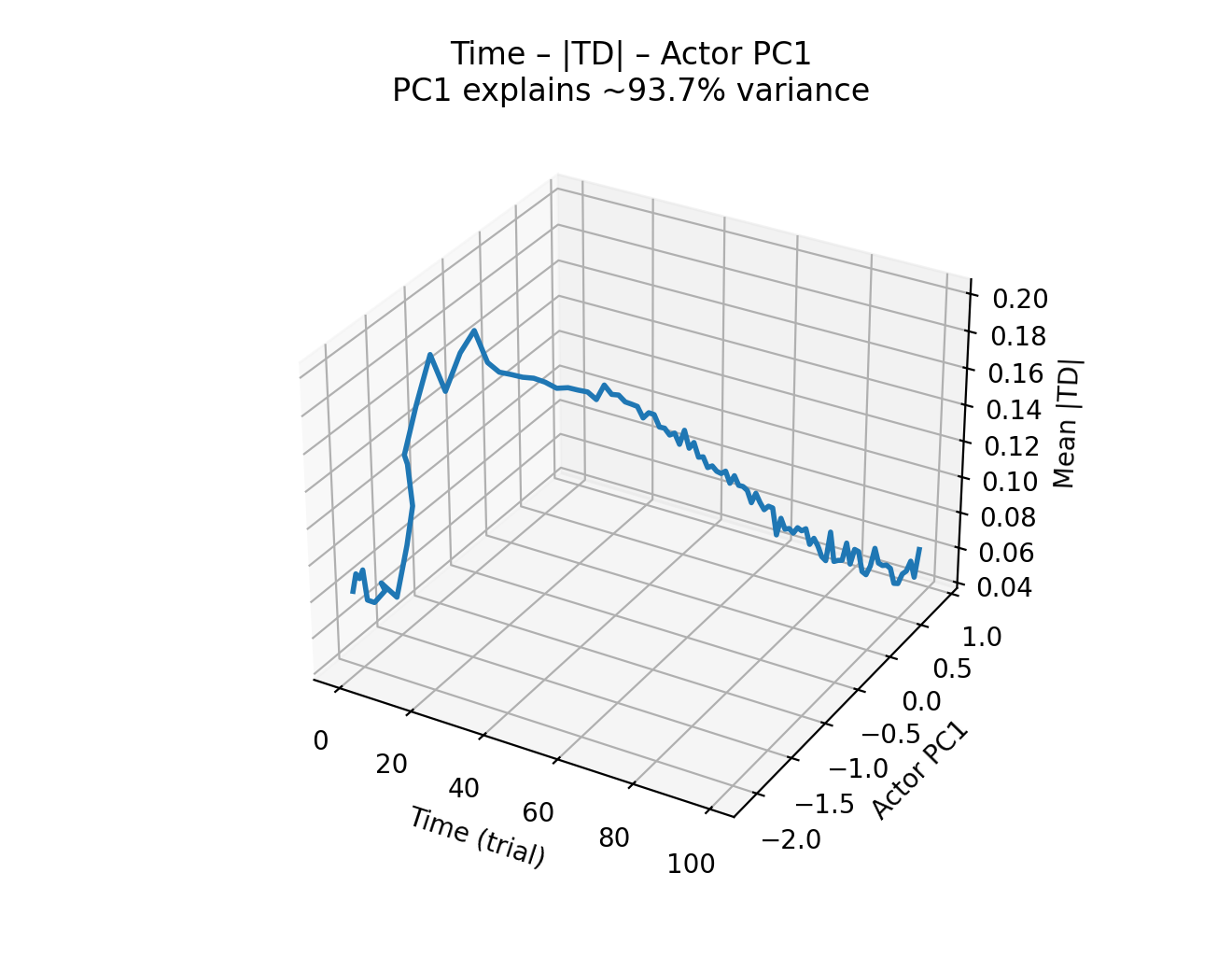}
         \caption{Actor weight with time and TD}
         \label{chapter4_fig:target_nosmoothingADHDPactweightTD}
     \end{subfigure}
     \hfill
     \begin{subfigure}[b]{0.5\textwidth}
         \centering
         \includegraphics[width=\textwidth]{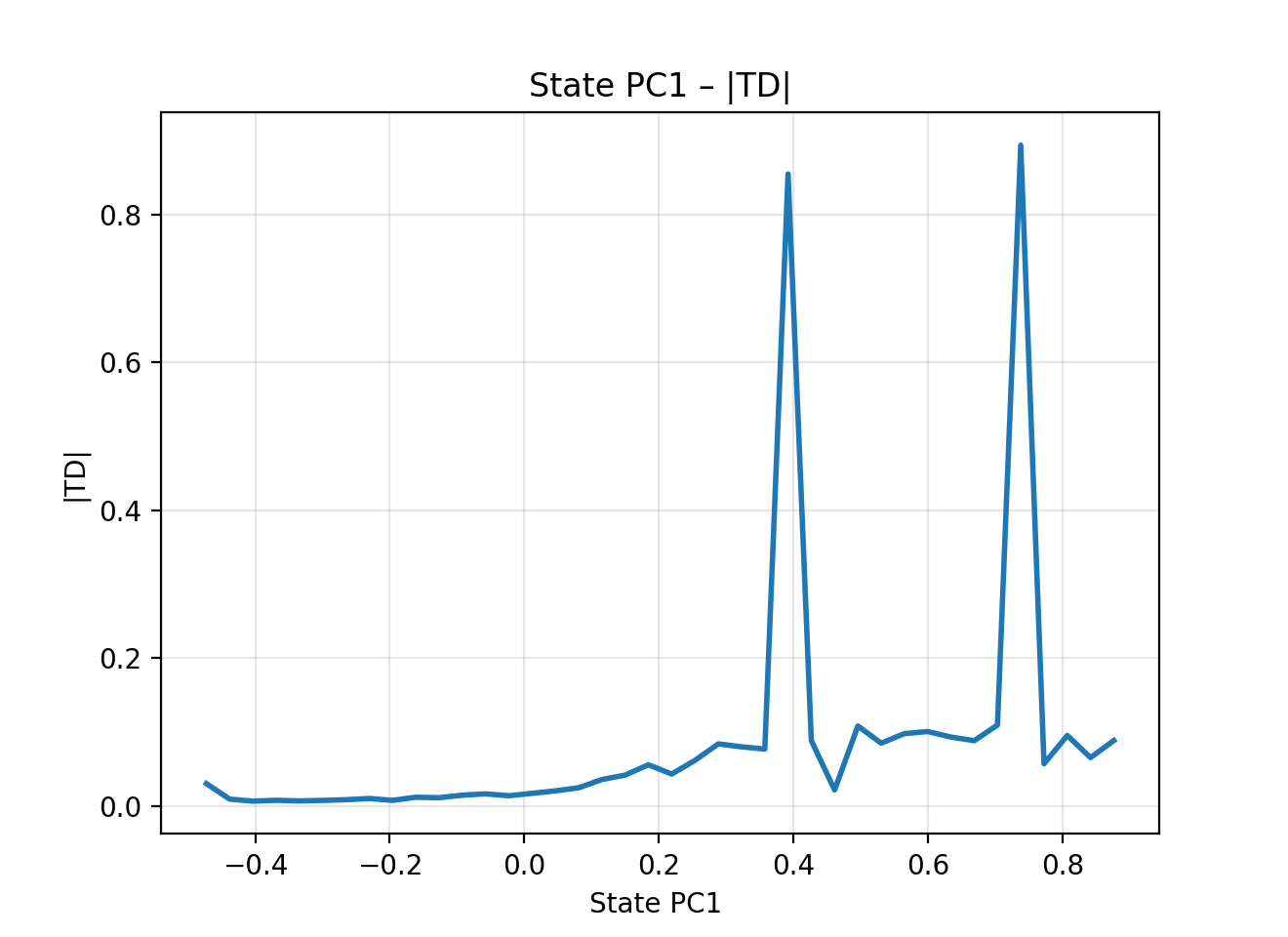}
         \caption{State with TD}
         \label{chapter4_fig:target_nosmoothingADHDPstateTD}
     \end{subfigure}
        \caption{TD trajectory, ADHDP with training stabilizers(Huber critic loss and numeric cost scaling), no target policy smoothing}
        \label{chapter4_fig:target_nosmoothingADHDPTD}
\end{figure}

In \autoref{chapter4_fig:target_nosmoothingADHDPTDbytrial}, it shows that the TD error varies within a relatively narrow range and gradually decreases to a lower value during training. This indicates that the Huber loss and cost scaling still help suppress large TD fluctuations. However, the loss landscapes reveal that this decrease of TD does not correspond to a good optimization region. As shown in \autoref{chapter4_fig:target_nosmoothingADHDPcritic}, the critic match loss landscape exhibits a very sharp and narrow peak, which is much steeper than the version with target policy smoothing. This sharp geometry is caused by the deterministic TD target without smoothing. The actor loss landscape in \autoref{chapter4_fig:target_nosmoothingADHDPactorloss} still exhibits a wedge-shaped valley, indicating that the policy gradient remains much stronger along one direction.

This behavior is also reflected in the training trajectory. In \autoref{chapter4_fig:target_nosmoothingADHDPactweightTD}, the TD decreases while the first principal direction of the actor weight gradually moves toward a single dominant direction. This shows that the actor update is still biased and does not sufficiently explore other directions of the policy space. The state–TD relation in \autoref{chapter4_fig:target_nosmoothingADHDPstateTD} still contains two clear peaks. Although the amplitude of these peaks is smaller than in the basic ADHDP and target network versions, they indicate that certain state regions still produce relatively large TD errors.

Under this biased update direction, the control policy is gradually pushed toward the torque saturation boundary. As shown in \autoref{chapter4_fig:target_nosmoothingADHDPstates}, the control torques remain close to the saturation limit during the last trial. Once the control inputs approach the limit, the controller cannot generate sufficient corrective torque to stabilize the spacecraft. As a result, the angular velocity and quaternion trajectories fail to converge. These observations explain why the TD curve appears smooth while the closed-loop control still diverges.

In conclusion, after removing target policy smoothing, the critic match loss returns to deterministic behavior, causing the spike in the critic match loss landscape to appear again. However, due to the presence of Huber, cost scaling, and small noise, the overall TD trend continues to decline. However, the policy updates guided by this are very stiff, causing the control torque to be trapped in a near-saturation state. These results clarify the role of target policy smoothing in the ADHDP framework. While the Huber loss and cost scaling mainly reduce TD magnitude and suppress extreme TD spikes, target policy smoothing primarily influences the geometry of the critic loss landscape by preventing the formation of sharp peaks. However, even with these stabilizers, the actor optimization remains strongly biased toward a dominant direction, which ultimately leads to saturation of the control torque and failure of the closed-loop control.

A comparison across the four ADHDP variants reveals a consistent mechanism underlying the failure of the closed-loop control. Although the introduction of target networks, Huber loss, cost scaling, and target policy smoothing significantly stabilizes critic learning and reduces TD fluctuations, these improvements do not fundamentally change the optimization geometry of the actor. In all variants, the actor loss landscape consistently exhibits a strongly anisotropic wedge-shaped structure, where the policy gradient is dominant along one direction while remaining weak along others. Under the torque saturation constraint, this geometry gradually drives the policy toward the action boundary. Once the control torques approach saturation, the controller loses the ability to generate effective corrective torques, and the spacecraft states eventually diverge. These observations suggest that the primary difficulty of the system does not lie solely in critic instability, but rather in the unfavorable optimization geometry arising from the coupling between actor and critic updates. Improvements in critic stability can smooth the TD signal and reduce short-term fluctuations, yet they do not fundamentally alter the dominant-direction structure of the actor loss landscape. Consequently, the learned policy still tends to drift toward the saturation boundary, ultimately leading to control failure.

The visualization results therefore suggest that further improvements should focus not only on stabilizing critic learning, but also on improving the geometry of the actor optimization landscape. Possible directions include introducing stronger regularization on control effort, adjusting the actor output scaling relative to the torque limits, or designing objective formulations that penalize near-saturation actions more explicitly. Such approaches may help reshape the actor loss landscape and prevent the policy from being trapped near the saturation boundary while preserving the lightweight online structure of the ADHDP framework.

From the results discussion above, the following summary can be drawn. The visualization framework provides a clear view of how different parts of the actor–critic algorithm evolve during training. The critic match loss landscape shows how the critic fits the TD targets and whether the critic optimization reaches a stable region in the parameter space. The actor loss landscape reveals the shape of the policy and whether the actor update follows a smooth direction. The time–TD–actor weight trajectory illustrates how the actor parameters change together with the TD signal, while the state–TD plot identifies which system states produce large TD errors. Together, these visualizations explain how the critic, actor, and system states interact during learning, and how their coupling influences training stability and control performance. Through these observations, the influence of different ADHDP modifications can be directly identified, demonstrating that the framework is useful for analyzing and comparing reinforcement learning algorithms. This indicates that the framework provides a practical tool for interpreting the training behavior of actor–critic reinforcement learning algorithms.

\section{Conclusion}

In this work, a loss landscape visualization framework is developed to interpret the training process of actor–critic reinforcement learning algorithms. The framework includes four visualization components. The critic loss landscape shows how the critic fits the TD targets during training. The actor loss landscape, with a frozen critic, reflects the shape and quality of the learned policy. The state–TD plot shows which state regions produce large TD errors, and the actor weight–time–TD plot tracks how the policy changes with the TD error. Four ADHDP variants are used to demonstrate the framework. The results show how training stabilizers and target updates change the loss landscape and affect training stability. The visualizations also show that a small TD error does not always mean a good policy. Overall, the proposed framework provides a systematic approach for diagnosing reinforcement learning algorithms and for understanding how algorithmic design choices influence training behavior and control performance.











\bibliographystyle{unsrt} 
\bibliography{main}


\end{document}